%% file: main.tex
\documentclass[10pt,twocolumn,letterpaper]{article}

\usepackage[pagenumbers]{iccv} 

\input{defs}

\definecolor{iccvblue}{rgb}{0.21,0.49,0.74}
\usepackage[pagebackref,breaklinks,colorlinks,allcolors=iccvblue]{hyperref}

\usepackage{graphicx}
\usepackage{array,multirow,multicol}
\usepackage{pgffor} 
\usepackage{pgfmath} 
\usepackage{caption}
\usepackage{subcaption}
\usepackage{adjustbox}

\def\paperID{1370} 
\def\confName{ICCV}
\def\confYear{2025}

\title{Counting Stacked Objects}

\author{
Corentin Dumery$^{\scriptsize 1}$, Noa Etté$^{\scriptsize 1}$, Aoxiang Fan$^{\scriptsize 1}$, Ren Li$^{\scriptsize 1}$, Jingyi Xu$^{2}$, Hieu Le$^{\scriptsize 1}$, Pascal Fua$^{\scriptsize 1}$\\
\\
$^{1}$EPFL, $^{2}$Stony Brook University\\
{\tt\small \{name.surname\}@\{epfl.ch, stonybrook.edu\}}
}

\begin{document}

\input{figs/teaser.tex}

\input{sec/0_abstract}    
\input{sec/1_intro}
\input{sec/2_related_works}
\input{sec/3_method}
\input{sec/4_experiments}

\input{sec/5_conclusion}

\section{Acknowledgements.}

We would like to sincerely thank all the participants who spent time to play our counting game and provided us with valuable data on human performance at this task. In particular, we would like to congratulate Martin Engilberge, who gave the most accurate estimates and has been awarded a box of chocolates. We would also like to thank Nicolas Talabot, Víctor Batlle and Adriano D'Alessandro for insightful discussions.
This work was supported in part by the Swiss National Science Foundation.

{
    \small
    \bibliographystyle{ieeenat_fullname}
    \bibliography{new,string,geom,graphics,learning,vision,biomed,cfd}
}

\input{sec/X_suppl}

\end{document}

%% file: defs.tex
\newif\ifdraft
\draftfalse

\definecolor{burntorange}{rgb}{0.8, 0.33, 0.0}
\definecolor{orange}{rgb}{1,0.5,0}
\definecolor{green0}{rgb}{0.1,0.7,0.1}

\ifdraft
\newcommand{\PF}[1]{{\color{red}{\bf PF: #1}}}
\newcommand{\pf}[1]{{\color{red} #1}}
\newcommand{\CD}[1]{{\color{blue}{\bf CD: #1}}}
\newcommand{\cd}[1]{{\color{blue} #1}}
\newcommand{\AF}[1]{{\color{green0}{\bf AF: #1}}}

\newcommand{\NE}[1]{{\color{violet}{\bf NE: #1}}}

\newcommand{\RL}[1]{{\color{orange}{\bf RL: #1}}}

\newcommand{\HL}[1]{{\color{burntorange}{\bf HL: #1}}}

\newcommand{\JX}[1]{{\color{brown}{\bf JX: #1}}}

\newcommand{\TODO}[1]{\textbf{\color{red}[TODO: #1]}}
\else
\newcommand{\PF}[1]{{\color{red}{}}}
\newcommand{\pf}[1]{ #1 }
\newcommand{\CD}[1]{{\color{blue}{}}}
\newcommand{\cd}[1]{ #1 }
\newcommand{\AF}[1]{{\color{green0}{}}}

\newcommand{\NE}[1]{{\color{violet}{}}}

\newcommand{\RL}[1]{{\color{orange}{}}}

\newcommand{\HL}[1]{{\color{pink}{}}}

\newcommand{\JX}[1]{{\color{brown}{}}}

 \newcommand{\TODO}[1]{}
 
\fi

\newcommand{\parag}[1]{\vspace{-2mm}\paragraph{#1}}

\newcommand{\acron}{{\it 3DC}}

%% file: figs/teaser.tex

\twocolumn[
{
\renewcommand\twocolumn[1][]{#1}%
\maketitle
\begin{center}
\vspace{0em}
\includegraphics[width=\textwidth]{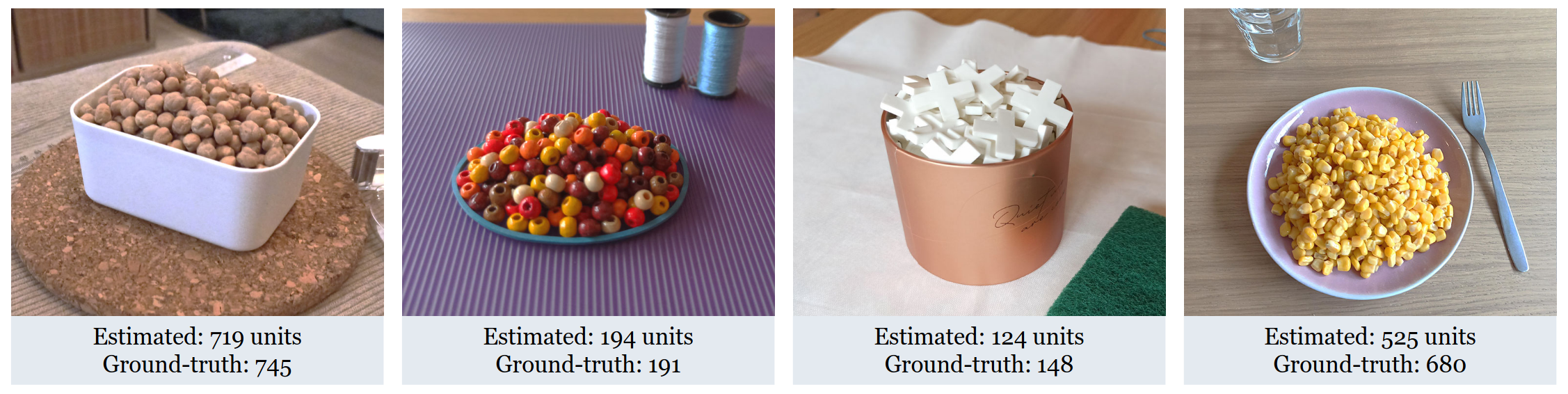}
\captionof{figure}{\textbf{3D Counting (3DC).} From multiple views of objects to be counted and their container, we estimate both the total volume they occupy and the fraction of this volume taken up by the objects. Combining these estimates yields the total number of objects.
}
\label{fig:teaser}
\end{center}
}]


%% file: sec/0_abstract.tex

\begin{abstract}

Visual object counting is a fundamental computer vision task underpinning numerous real-world applications, from cell counting in biomedicine to traffic and wildlife monitoring. However, existing methods struggle to handle the challenge of stacked 3D objects in which most objects are hidden by those above them.  To address this important yet underexplored problem, we propose a novel 3D counting approach that decomposes the task into two complementary subproblems - estimating the 3D geometry of the object stack and the occupancy ratio from multi-view images. By combining geometric reconstruction and deep learning-based depth analysis, our method can accurately count identical objects within containers, even when they are irregularly stacked. We validate our 3D Counting pipeline on large-scale synthetic and diverse real-world datasets with manually verified total counts. 
Our datasets and code and can be found at \small{\url{https://corentindumery.github.io/projects/stacks.html}}
\end{abstract}

\vspace{-1em}

%% file: sec/1_intro.tex

\section{Introduction}
\label{sec:intro}

Visual object counting---the task of quantifying the number of instances in a scene---serves as a fundamental building block for numerous real-world applications and autonomous decision-making systems. This ranges from cell counting in biomedical imaging~\cite{Xie18b} to traffic~\cite{Mandal2020} and wildlife~\cite{Arteta16}  monitoring. However, these methods~\cite{Liu22a, Xu23d, Ranjan21, Shi22a,Lu18} can only count visible objects such as apples spread across a table or people in a crowd. The problem becomes significantly harder when objects are stacked on top of each other, as in Fig.~\ref{fig:teaser}: Only a subset of them is visible, making counting much more difficult. In fact, our experiments show that this task is truly challenging even for humans. Nevertheless, solving it would have significant applications in industrial and agricultural settings, where precise quantification of items—such as products on a pallet or fruits in crates—not only prevents stock and quality errors but also enhances operational efficiency and logistics.

\input{figs/pipeline}

Overcoming the above-mentioned challenges requires inferring the presence and quantity of hidden instances from limited visual cues. This means not only detecting visible object features but also reasoning about hidden ones through contextual understanding. The challenge is further amplified by variations in stacking patterns, object orientations, and irregular arrangements, making traditional counting approaches inadequate. 

At the heart of our proposed solution is a key insight: the fraction of space occupied by objects, which we will refer to as the {\it occupancy ratio}, can be accurately inferred from a depth map computed by a monocular depth estimator from a view in which enough objects of interest are clearly visible.  In most cases, such a view is one where the container is seen roughly from above, without having to be strictly vertical.  To exploit it, we  break down the problem into two complementary tasks: estimating the 3D geometry of the object stack and estimating the occupancy ratio within this volume, as depicted by Fig.~\ref{fig:pipeline}. This decomposition enables us to solve the 3D counting problem through a combination of geometric reconstruction for volume estimation and deep learning-based depth analysis for occupancy prediction, both of which can be solved efficiently. 

We validate \acron{} through extensive experiments on real-world and synthetic datasets. Our real-world evaluation leverages a diverse collection of scenes depicting objects stacked in containers or still in their packaging, as shown in \cref{fig:teaser}. To further assess the reliability of 3DC, we also constructed a large-scale synthetic dataset with precisely annotated ground-truth counts. This dataset, along with our code, will be made available upon publication. 

Thus, our contributions are:
\begin{itemize}
 \item A complete pipeline for 3D counting of overlapping, stacked objects, a novel and challenging computer vision task not previously addressed in the literature.
 \item A network designed to infer the occupancy ratio, which embodies a novel idea and forms a key component of the architecture. 
 \item An extensive new 3D Counting Dataset comprising 400,000 images from 14000 physically simulated and rendered scenes with precise ground-truth object counts and volume occupancy computed programmatically.
 \item A complementary real-world validation dataset consisting of 2381 images from 45 scenes captured with accurate camera poses and manually verified total counts.
 \item A human baseline derived from 1485 annotations on real images, representing estimates from 33 participants. 
\end{itemize}
In particular, the latter shows that this task is truly hard and that humans perform poorly. This indicates that training a network to predict stacked counts from images  in a single step may not be feasible and that our decomposing the problem into simpler subproblems is critical to success, as demonstrated in~\cref{sec:experiments}.

%% file: figs/pipeline.tex

\begin{figure*}[htbp]
    \centering
    \includegraphics[width=\linewidth]{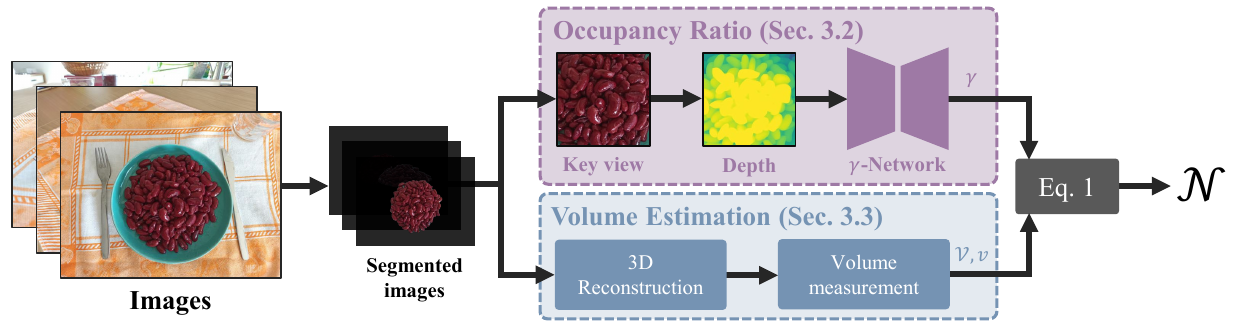}
    \vspace{-8mm}
    \caption{\textbf{3DC pipeline.} We decompose the counting task into estimating the volume of the objects to be counted and then estimating the occupancy ratio within that volume. The first is done on the basis of geometry reconstructed from segmentations in multiple images.The second uses as input a depth-map computed by a monocular depth estimator and regresses an occupancy ratio from it. }
    \label{fig:pipeline}
\end{figure*}

%% file: sec/2_related_works.tex

\section{Related work}
\label{sec:rw}

Counting aims to estimate the number of instances of a specific object category in a scene. Most methods focus on counting visible objects from a single image. A few methods leverage multiple images to enhance counting accuracy. Our method extends this to the more challenging scenario where many of the objects to be counted are hidden. We summarize those related work below.

\parag{Single-View Counting.} Most recent counting methods focus on single-view scenarios where they train a network specialized in counting a single object-category such as for crowd-counting~\cite{Xu19a,Shi19b,Liu19d,Sindagi19a,Cheng19a,Wan19b,Zhang19b,Zhang19c}, counting cars~\cite{Mandal2020}, or penguins~\cite{Arteta16}. These methods are widely applicable~\cite{Chattopadhyay16}, such as when counting cells or other anatomical structures in medical imaging~\cite{Falk18} or counting trees or building in satellite images~\cite{Yi23}. The proposed algorithms address challenges such as scale variation, perspective distortions, and occlusions. Common approaches are to learn robust feature representations ~\cite{Zhang16c}, to perform density map estimation~\cite{Sindagi17}, or to leverage multi-scale features~\cite{Ranjan18}. Apart from the traditional setting in which objects of a single category are counted, class-agnostic counting~\cite{Lu18,Nguyen22a} enables counting an arbitrary category at test time, given a few image samples of the class~\cite{Yang21f}, a few bounding boxes~\cite{Ranjan21}, or just the class name~\cite{Xu23d}. Like class-agnostic counting methods, our approach is not restricted to specific categories but can generalize to any object type at test time. For the most part, counting methods only deal with visible objects. The only related attempt to ours is the approach of~\cite{Jenkins23} that infers counts of occluded objects by using LiDAR data. However, their algorithm only handles a specific setup of counting different beverages on shelves for a specific set of categories with known volumes, such as ``\textit{Coca-Cola 20oz bottle}'' or ``\textit{milk carton}'', while our method does not require LiDAR and generalizes to diverse scenes, object types, and geometries.

\parag{Multi-View Counting.} Multi-view counting approaches improve accuracy by combining information across multiple camera views, often projecting feature maps onto a common ground plane to generate precise density maps for crowd counting~\cite{Zhang20g,Zhang20b} or segmentation maps of fruits~\cite{Nellithimaru19}. However, these methods also assume that all objects are visible from at least one view and are restricted to a specific class of objects, making them inadequate in many practical scenarios. In contrast, our method targets the largely unexplored area of counting occluded objects, without any restriction on the nature and shapes of these objects. 

%% file: sec/3_method.tex

\section{Method}
\label{sec:method}

\acron{}  aims to estimate the total number $\mathcal{N}$ of objects in a container solely from a set of 2D images,  which is a challenge even for humans and has not been attempted before as far as we know. Our approach is predicated on the idea that, even though we cannot faithfully recover the exact arrangement of all invisible objects at the bottom of the container, the occupancy ratio can be estimated from a single image, provided that enough objects are visible in it. 

To localize the objects in cluttered scenes that may contain several stacks, we rely on a segmentation of the objects $S_{o}$ and their container $S_{c}$ in the first frame. We then use SAM2~\cite{Ravi24a} to propagate this segmentation as $S_{o,f}$ and $S_{c,f}$ to all subsequent frame $f$, so that  the frame in which the most objects are visible can be automatically identified.

In the remainder of this section, we first formalize our approach. We then introduce our occupancy ratio estimator, followed by our approach to estimating volumes.

\subsection{Problem Statement}
\label{sec:statement}


Assuming the average volume occupied by a single object is $v$, and the total volume of the container is $\mathcal{V}$, it would be tempting to compute the number of objects as $\mathcal{N} = \frac{\mathcal{V}}{v}$. However, this fails to account for gaps between objects. If we consider that the objects are stacked in such a way that only a fraction $\gamma$ of the volume $\mathcal{V}$ is actually taken up by the objects themselves and that the rest is empty space, then the previous estimate becomes
\begin{align}
    \mathcal{N} = \frac{\mathcal{\gamma V}}{v} \label{eq:counting}
\end{align}
Our key insight is that this volume usage rate $\gamma$ over the whole container can be estimated with high accuracy from the visible elements only. In practice, the density within the container may not be strictly uniform. However, if the container is large enough, the variations tend to compensate each other over the whole volume and using an average value is warranted. 

\parag{Assumptions.}

In this work we assume that the objects are stacked uniformly in bulk and approximately identical. Some objects are expected to be partially visible, so that the occupancy ratio $\gamma$ can be estimated. 

\parag{Applicability.}

The above assumptions hold in the real-world scenes of ~\cref{fig:teaser}. Furthermore, they are weak enough to also hold in many realistic scenarios across various industries. 
In warehousing and retail, our proposed setup can be used to automate the inventory process by accurately counting stacked items, reducing the need for manual labor and easing restocking. 
In manufacturing, our method can enhance quality control by ensuring that shipped containers include a sufficient number of items. 
It can also provide 3D scene understanding to autonomous systems for robotic tasks like pick-and-place and sorting. 

\input{figs/samples}

\subsection{Occupancy Ratio Estimation}
\label{sec:occupancy}

The most critical step in our approach is estimating the occupancy volume ratio $\gamma$ of Eq.~\ref{eq:counting} from a single image in which enough target objects can be seen, typically one taken from above  even though this is not a strict requirement. 
 
Formally,  we seek to learn a function $\Phi: \mathcal{D} \rightarrow \gamma \in \ensuremath{[0,1]}$ that takes as input a depth map and predicts an occupancy ratio $\gamma$. $\Phi$ learns the relationship between depth maps and occupancy ratio, capturing a key fact: if objects deeper in the stack remain visible, then the gaps between objects are large and the percentage of the volume occupied by them is low.  This does not depend on the exact shape of the objects being observed, and should, in theory,  apply even to new inputs with shapes not seen during training. In the results section, we will confirm this to be true in practice.

\parag{Network Architecture.}

To implement $\Phi$,  we use an encoder-decoder architecture that first computes rich image features and then decodes them into our target $\gamma$. For the encoder, we use DinoV2~\cite{Oquab23}, a foundation model trained on many real-world images to help with generalizability while increasing convergence speed. 


The decoder has to aggregate feature values into a single scalar representing the volume usage, predicted for the whole image. To this end, we use consecutive convolutional layers to gradually decrease the resolution of the features, along with ReLU activations, reducing the encoded feature image to a single pixel with 64 channels, and a final linear layer to predict a single scalar from the output of the last convolutional layer. We refer the reader to our supplementary material for additional details.

\parag{Training Data.}  

To train $\Phi$, the simplest is to minimize the squared error between the estimated occupancy ratio and ground-truth one over an annotated dataset. Unfortunately, no such dataset exists and we therefore synthesized our own. It comprises 400,000 images spanning 14,000 scenes containing various objects in different containers.  Fig.~\ref{fig:samples} depicts some of them. 

To create it, we used the ABC dataset~\cite{Koch19a} that features a wide variety of \textit{computer-assisted design} (CAD) models. We retained only watertight objects with a single connected component and rescaled them to fit in a cube of side $0.05$.
We then generated a virtual 3D scene with a container, and used a physics-based simulator to drop an initial batch of 100 identical objects in that box. We repeated this step a random number of times or until the box was full, that is to say, the union of objects reached the space above the box after the physical simulation has converged. In each scene, the container is given a random shape and scale. We also include some scenes without any container where objects are directly stacked on the floor, and some scenes where boxes are partially full, as is often the case inreal-life. 

Once the physical simulation was complete, we computed analytically the occupancy ratio as well as the total number of objects in the container. To each object and container, we randomly assigned a realistic material that could be metallic, wood, or plastic. In some scenes, we selected a different material for each individual object. Finally, we used a ray-tracing engine to render multiple realistic images of the container and objects seen from several different angles to allow 3D reconstruction, as shown  in \cref{fig:multiview}. Crucially, realistic rendering causes objects at the bottom of the boxes to appear darker due ambient occlusion. 

We repeated this for over 14,000 scenes and set aside a subset of 100 to use as a test set of shapes unseen during training. For each one of the 4800 shapes used, we generate a scene with a unit-cube container to reliably measure the ground truth $\gamma$ as well as two additional scenes with random containers.  This dataset allows us not only to train $\Phi$ on the top view, but also to run our complete pipeline on the multi-view images in order to measure the accuracy of our count estimate $\mathcal{N}_{est}$, as performed in \cref{sec:experiments}. \cd{We train our model on depth maps produced by the depth estimator, but} our dataset also includes ground-truth depth maps, which we employ in an ablation study in our experiments to assess the requirement of accurate depth maps of our method. We report additional statistics that highlight the diversity of the proposed dataset in \cref{fig:dataset_stats} of our supplementary.

\begin{figure}[t]
    \centering

    \begin{subfigure}{0.32\columnwidth}
        \centering
        \includegraphics[width=\textwidth]{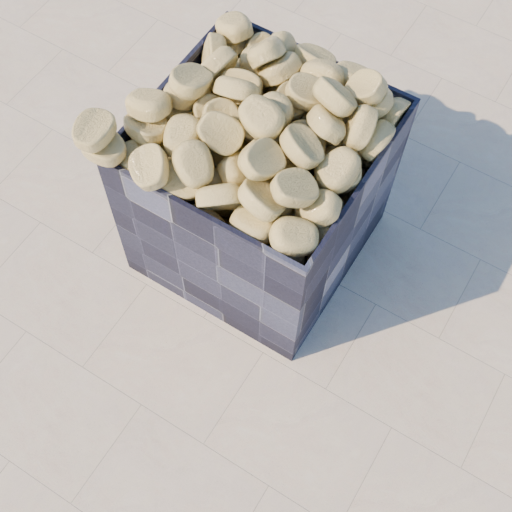}
    \end{subfigure}
    \hfill
    \begin{subfigure}{0.32\columnwidth}
        \centering
        \includegraphics[width=\textwidth]{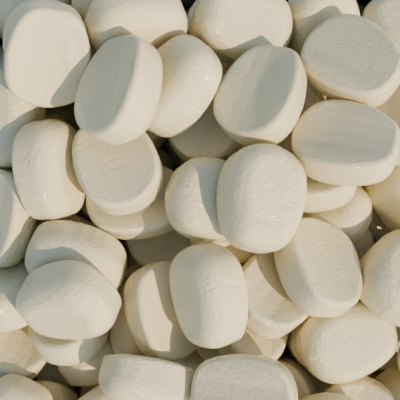}
    \end{subfigure}
    \hfill
    \begin{subfigure}{0.32\columnwidth}
        \centering
        \includegraphics[width=\textwidth]{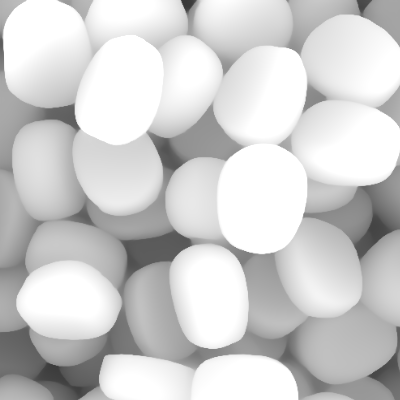}
    \end{subfigure}

    \vspace{0.3em}

    \begin{subfigure}{0.32\columnwidth}
        \centering
        \includegraphics[width=\textwidth]{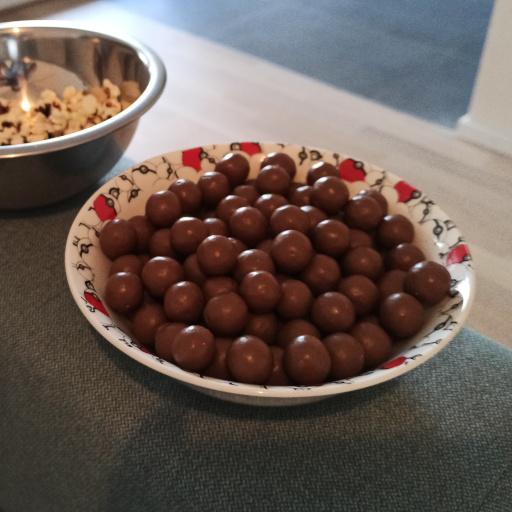}
        \subcaption*{(a)}
    \end{subfigure}
    \hfill
    \begin{subfigure}{0.32\columnwidth}
        \centering
        \includegraphics[width=\textwidth]{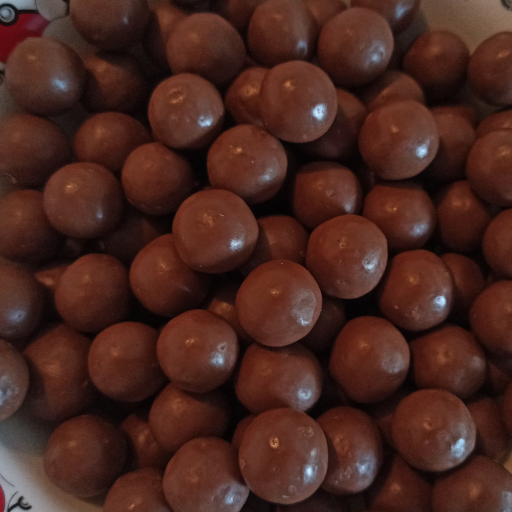}
        \subcaption*{(b)}
    \end{subfigure}
    \hfill
    \begin{subfigure}{0.32\columnwidth}
        \centering
        \includegraphics[width=\textwidth]{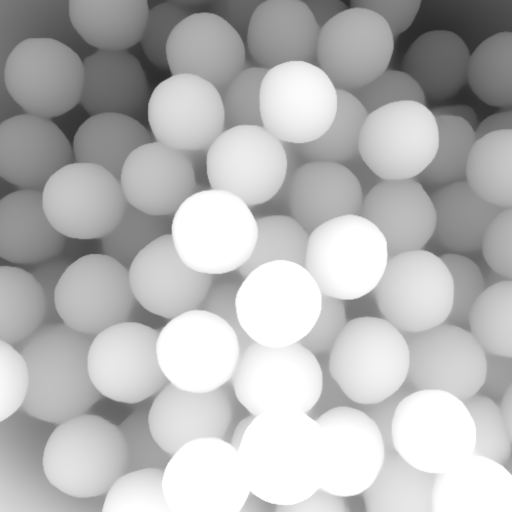}
        \subcaption*{(c)}
    \end{subfigure}

    \vspace{-0.3em}

    \caption{\textbf{Reducing the domain gap.} Instead of estimating the occupancy ratio $\gamma$ from synthetic (top) and real images (bottom) (a), we identify a \textit{key view} (b) and train a network to predict $\gamma$ from their depth maps (c), which are indistinguishable. 
    Top row: synthetic, $\gamma_{gt} = 62.4\%$. Bottom row: \textit{chocolates}, $\gamma_{est} = 53.5\%$, $\mathcal{N}_{est} = 119$, $\mathcal{N}_{gt} = 131$.}
    \label{fig:similarity}
\end{figure}

\parag{Inference.} On real-world data, it remains to determine which image to use as input to the $\gamma$-network. We will refer to this image as the \textit{key view}. To produce the best results, this view should be as close as possible to the depth maps seen during training, as illustraed in ~\cref{fig:similarity}. We automatically select the view that has the largest object segmentation, ensuring the objects are clearly visible, and crop the image to include only the masked content. We then employ Depth Anything V2~\cite{Yang24c} to compute a depth map. Note that we also train $\Phi$ on depth maps predicted by that model instead of the ground-truth depth maps, which further reduces the domain gap as demonstrated in our experiments.

\subsection{Volume Estimation}
\label{sec:volume_estimation}

Given an estimate of the occupancy ratio $\gamma$ obtained as discussed above, we still need to compute the total volume of the stack $\mathcal{V}$ to derive the total number of objects from Eq.~\ref{eq:counting}, and the unit volume $v$ if it is not known. When the cameras are uncalibrated, we use COLMAP~\cite{Schoenberger16a} to compute their poses and adjust their scale using a real-world reference measurement. \pf{In industrial scenarios with fixed camera setups known {\it a priori}, simpler methods could be be used.}

Inferring the volume $\mathcal{V}_{est}$ of the stacked objects from multiple images is a well-understood problem. In our specific implementation, we start from the segmented images and extract the container and objects by adding the mask as alpha channel. We then optimize 3D Gaussian Splats~\cite{Kerbl23} from these images, ensuring our reconstruction covers the container and objects only. To compute the volume the splats enclose,  we adapt the voxel carving algorithm. We initialize a voxel grid from the bounding box of the gaussians. Then, given the masks generated previously and depth maps rendered from 3DGS, a voxel is carved out if its projection on any given view falls outside of the mask or if its projected depth is less than the reconstructed depth map. This procedure successfully reconstructs objects and their containers, including partially filled boxes.

\input{figs/multiview}

To remove the container from the reconstruction, we estimate its thickness $t$ and erode the voxels on all sides except the top by $t$, thereby refining the estimated volume to represent only the contents. The value of $t$ is predicted by an additional decoder $\Psi$ that takes the same encoded image features as our previous network $\Phi$ in~\cref{sec:occupancy}. To this end, we use dilated convolutional layers~\cite{Yu15}, increasing the network's receptive field with no additional parameters. To make this prediction scale independent and easily predictable from 2D images alone, we predict thickness as a ratio of the container's size. We supervise $\Psi$ with ground-truth thickness from our dataset and at inference we average the estimation of $\Psi$ over all images.
Taken together, these algorithms yield a good estimate of the volume $\mathcal{V}$ spanned by the stacked objects.

\parag{Unit volume $v$.}
In most applications, the unit volume $v$ of Eq.~\ref{eq:counting} is known exactly because the object has been manufactured to a precise specification, $v$ can be computed from the simple geometry of the objects, or $v$ can be obtained from existing reference data, particularly for food items. When the unit volume of an object $v$ is not readily available, we estimate its value using the method described above and from a set of images of a template object. This task is made easier by the absence of a container, and this is shape but not scene specific. For a new shape, one exemplar is selected in a single frame and we can then use SAM2~\cite{Ravi24a} to generate a segmentation on all frames at once. The unit volume $v$ computed this way can then be used across all scenes containing this object.

%% file: figs/samples.tex
\newcommand{\inputnames}{00006132,00009652,00003960,00008958,00008316,00009039}
\newcommand{\labels}{{2.9, 18.9, 29.4, 38.8, 49.3, 63.7}}

\begin{figure*}[ht]
    \centering
    \begin{minipage}[c]{0.01\textwidth}
        \centering
        \rotatebox{90}{\textbf{Render}}
    \end{minipage} \hspace{-0.27em}
    \foreach \input [count=\i from 0] in \inputnames {
        \begin{minipage}[c]{0.155\textwidth} 
            \centering
            \includegraphics[width=\textwidth]{images/samples/\input_RGB.png}
        \end{minipage}%
    }

    \vspace{0.15em}
    \begin{minipage}[c]{0.01\textwidth}
        \centering
        \rotatebox{90}{\hspace{1.5em}\textbf{Depth}}
    \end{minipage}
    \foreach \input [count=\i from 0] in \inputnames {
        \begin{minipage}[c]{0.155\textwidth} 
            \centering
            \includegraphics[width=\textwidth]{images/samples/\input_DepthGT.png} \\
            $\gamma = $\pgfmathparse{\labels[\i]}\pgfmathresult \% 
        \end{minipage}%
    }
    
    \caption{\textbf{Dataset samples.} We visualize generated scenes in ascending order of occupancy ratio, with ground-truth depth maps.}
    \label{fig:samples}
\end{figure*}

%% file: figs/multiview.tex
\begin{figure}[t]
    \centering

    \foreach \img in {1_0, 1_1, 1_2} {
        \begin{subfigure}{0.3\columnwidth}
            \centering
            \includegraphics[width=\linewidth]{images/multiview/\img.png}
        \end{subfigure}
    }
    \vspace{0.4em} 
    
    \foreach \img in {3_0, 3_1, 3_2} {
        \begin{subfigure}{0.3\columnwidth}
            \centering
            \includegraphics[width=\linewidth]{images/multiview/\img.png}
        \end{subfigure}
    }
    \vspace{0.4em} 

    \foreach \img in {2_0, 2_1, 2_2} {
        \begin{subfigure}{0.3\columnwidth}
            \centering
            \includegraphics[width=\linewidth]{images/multiview/\img.png}
        \end{subfigure}
    }
    \vspace{0.4em} 

    \hspace{-0.6em}
    \foreach \img in {4_0, 4_1, 4_2} {
        \begin{subfigure}{0.3\columnwidth}
            \centering
            \includegraphics[width=\linewidth]{images/multiview/\img.png}
        \end{subfigure}
    }
    
    \vspace{-0.4em} 

    \caption{\textbf{Multi-view images.} We generate 30 views from arbitrary angles for each of the scenes in our large-scale synthetic dataset.}
    \label{fig:multiview}
\end{figure}

%% file: sec/4_experiments.tex

\section{Experiments}
\label{sec:experiments}

We evaluate our method in two ways: measuring the accuracy of 3D counting as a whole and of the occupancy ratio estimation, in \cref{sec:eval_count} and \cref{sec:eval_vol}, respectively. These evaluations are performed over two datasets. The first comprises 100 scenes representing physically simulated shapes from the ABC dataset~\cite{Koch19a}. These scenes were isolated after their generation and were not seen during the training of our occupancy ratio network. The second is made of 2381 real images spanning 45 real scenes that were captured with a regular smartphone's RGB camera, and no additional sensor. These captures offer multiple views around stacks of objects in a container, lying flat on a table or still enclosed in their packaging. 
We count the ground-truth number of units manually for all scenes below 500, or infer it from the weight for even larger counts. In~\cref{fig:intermediate_results}, we provide intermediate results to help our readers form a better intuition about our method's behavior, and additional qualitative results in~\cref{fig:additional} and~\cref{fig:teaser}. 

\input{figs/results_table}

\subsection{Metrics}

We use several metrics to assess the accuracy of object counting and occupancy ratio estimation. The object counts vary significantly across scenes, ranging from 19 to 20063. Thus, the Mean Absolute Error (MAE)---defined as $\text{MAE} = \frac{1}{n} \sum_{i=1}^n |y_i - \hat{y}_i|,$ where \( y_i \) is the ground truth count and \( \hat{y}_i \) is the predicted count for each scene---tends to amplify the importance of scenes with a high counts. To mitigate this, we also  report normalized metrics. We use the Normalized Absolute Error (NAE), Squared Relative Error (SRE), and Symmetric Mean Absolute Percentage Error (sMAPE), which scale errors relative to the ground truth. The NAE provides a measure of the absolute error normalized by the total ground truth count across scenes, SRE emphasizes larger errors and penalizes significant deviations in high-count scenes, and , sMAPE offers a normalized percentage error. The exact formulas for all metrics can be found in our supplementary.

\subsection{Counting Evaluation}
\label{sec:eval_count}

\input{figs/additional}



\input{figs/intermediate}

As far as we know, there is no previous work on counting from multiple images which does not either implicitly assume all objects to be visible or require additional sensors such as LiDARs. Thus, we compare \acron{} against BMNet+~\cite{Shi22a}. It predicts a density map over all pixels of an image, and the estimated count is then inferred by summing over all pixels. Additionally, we compare against a combination of SAM~\cite{Kirillov23} and CLIP~\cite{Radford21}, where SAM is used to generate a large number of masks from an input image and CLIP uses a set of negative and positive text prompts to identify masks that represent an object of interest. The final count is then taken to be the number of these masks. In Tables~\ref{tab:eval_counting} and~\ref{tab:eval_counting_real}, we compare the performance of these two baselines against that of our method. We outperform them in both cases.  

In early experiments, we attempted to directly predict object count from images. To this end, we trained different networks, coined \textit{ViT+H} and \textit{CNN}. We also report the results in Tables~\ref{tab:eval_counting} and~\ref{tab:eval_counting_real}. They perform poorly, especially on the real-world dataset, which is what prompted us to look into decomposing the problem into occupancy and volume estimation. 

We also sought to estimate how good humans are at this counting task. To this end, we organized a contest and encouraged participants to make accurate guesses on the 45 real scenes. The contest registered 1485 guesses from 33 participants. We define the \textit{Human} baseline as the average of error metrics of participants, and \textit{Human-Vote} as the error of the average guess across all participants. This second baseline should be stronger as the errors of participants tend to cancel-out and that is what we observe in Tab.~\ref{tab:eval_counting_real}. However, the results are still much worse than what our approach delivers. Interestingly, participants who spent a longer time did not perform better than their peers, highlighting the difficulty of this task. \cd{When asked about the method they used, a majority of participants reported counting the number of objects on each axis, and multiplying these values together. This proved to be ineffective, however.} 

Finally, we tested  \textit{LlamaVision 3.2 11B}, a Large Language and Vision Model, on our counting task. While it was very good at describing the physical appearance of the objects in the real scenes and the composition of the scene, its count estimates were completely off as shown in Tab.~\ref{tab:eval_counting_real}.

Our approach beats all the aforementioned baselines, providing the first method to estimate stack counts with reasonable accuracy. However, we also note that our perfomance remains better on the synthetic data than on our benchmark of real scenes. This can be attributed to the additional complexity of these real scenes, which often contain thousands of objects and can be more challenging for volume estimation.

\subsection{Occupancy Ratio Evaluation}\label{sec:eval_vol}

We now focus on the occupancy ratio network alone. There is  little work on occupancy ratio estimation, we thus implemented additional baselines to gain insights about the inner working of our approach. 

Since we assume that the depth map contains enough information to predict the occupancy ratio, we define a first baseline that we will refer to as \textit{DepthExtrapolated}. Given the top view of the container, we compute the maximal depth using a monocular depth estimator and use it to normalize the depth map. Then, we average the resulting values of the $K$ pixels, yielding the volume fraction estimate
$$
\gamma_{est}^{norm} = \frac{1}{K} \sum \frac{d_i}{d_{max}} \; .
$$
This first baseline tends to predict values lower than expected.  Thus, we defined a second one we dubbed  \textit{DepthCorrected}. It uses linear regression to correct $\gamma_{est}^{norm}$ into a new estimate $\gamma_{est}^{corrected}$. This yields a method able to model the observation that depth maps with high variance tend to correspond to low occupancy ratios. Finally, we also compare with the mean estimator, that predicts the mean percentage of $32.3\%$ occupancy ratio for all inputs.

As can be seen  in \cref{tab:eval_volume}, our method outperforms these three baselines by a significant margin. \textit{DepthCorrected} is better than the other two baselines, which shows that depth information is indeed useful for this task.  However, it does not fully predict the occupancy ratio. We interpret this as a strong clue that the ratio $\frac{d_i}{d_{max}}$ alone is not enough to predict $\gamma$, and our proposed network successfully learns to extract meaningful information from the depth maps. We hypothesize that our volume network captures additional geometric information such as the influence of concavities in the final occupation of volume.

\subsection{Ablation Study}

\input{figs/ablation_table}

We ran additional experiments to evaluate the sensitivity of our approach to the depth maps produced by the monocular depth estimator. 

Since our synthetic dataset has ground-truth depth maps both for the training images and the validation images, we use them as follows. Recall from \cref{sec:occupancy} that, at training time, we use Depth Anything V2~\cite{Yang24c} depth maps, a setting we refer to as $\mathcal{T}-$ for training without ground-truth. Instead, we could use ground-truth depth maps during training, a setting we will refer to as $\mathcal{T}+$. Similarly, at inference time, we can use the estimated depth map. which is what we normally do, or the ground-truth one. We refer these as $\mathcal{D}-$ and $\mathcal{D}+$, respectively. Thus, the standard  configuration of our method is $\mathcal{T}-,\mathcal{D}-$. The others are only used for ablation purposes. 

We report the results in \cref{tab:ablation_counting} and occupancy ratio estimation in \cref{tab:ablation_volume}. Entirely dropping the ground-truth depth maps and relying {\it only} on estimated ones, which is our standard operating procedure, is best. We hypothesize that this is due to the slight smoothing in the produced depth maps, which may prevent the model from overfitting to specific shape features in perfect depth maps.

Even though the network is trained on synthetic data, this observation further confirms the generalizability of \acron{}  to real data, since in a practical scenario such as the real data of \cref{tab:eval_counting_real}, ground-truth depth maps are not available.

\cd{
Finally, we also report in~\cref{tab:eval_counting_real} an ablated method \textit{Ours (Color)} where the $\gamma$-network takes an RGB image as input instead of a depth map. While this method still outperforms humans, it suffers from a significant performance drop and justifies the use of depth maps in our final approach.
}

\subsection{Limitations and Future Work}

\begin{figure}[t]
    \centering
    \includegraphics[width=0.95\linewidth]{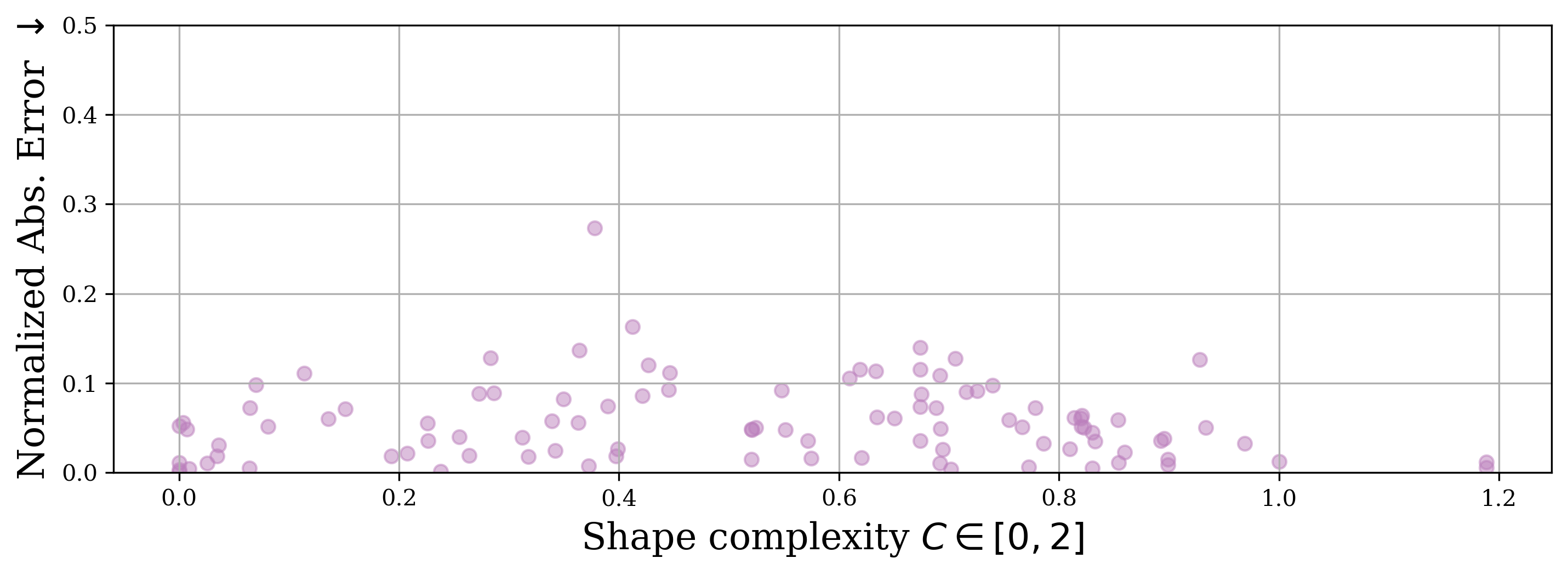}\vspace{-0.6em}
    \caption{\textbf{Complexity Analysis.} Each point represents the $\gamma$ occupancy ratio error for a shape in the validation set. 
    }\vspace{-0.6em}
    \label{fig:complexity}
\end{figure}
  
To evaluate the robustness of our method to complex shapes, we visualize in~\cref{fig:complexity} the error in occupancy ratio estimation as a function of shape complexity. The latter is measured by summing a curvature complexity term with the ratio of the shape's volume by the volume of its convex hull: 
$$
C = \frac{\kappa}{\|x_{max} - x_{min}\|_2 \kappa_0} + \frac{V_{hull}-V}{V_{hull}}
$$
where $\kappa$ is the integrated mean curvature of the shape, $\|x_{max} - x_{min}\|_2$ is a scaling factor and $\kappa_0$ is the maximum scaled $\kappa$ observed in the dataset. We only observe a slight error increase as the shapes become more complex, 

Unlike ours, many earlier methods  attempt to localize the objects being counted, increasing the interpretability and usability of the results. However, these localizations are often erroneous when objects are stacked together, as illustrated in \cref{fig:limitations}, greatly limiting their applicability. Thus another possible direction for further enhancements lies in integrating a robust localization of visible instances, and an estimation of a possible configuration of invisible instances.

\input{figs/limitations}

%% file: figs/results_table.tex

\begin{table}
    \centering
    \hspace{2em}
 \begin{small} 
    \begin{tabular}{l r r r r}
        \toprule
        & NAE $\scriptscriptstyle\downarrow$ & SRE $\scriptscriptstyle\downarrow$ & MAE $\scriptscriptstyle\downarrow$ & sMAPE $\scriptscriptstyle\downarrow$ \\
        \midrule
        BMNet+~\cite{Shi22a} & 0.91 & 0.87 & 320.50 & 158.87  \\
        SAM+CLIP~\cite{Kirillov23,Radford21} & 0.73 & 0.61 & 259.22 & 102.77 \\
        CNN & 0.66 & 0.48 & 235.74 & 98.44  \\
        ViT+H & 0.42 & 0.24 & 149.90 & 47.36  \\
        \hline
        Ours  & \textbf{0.22} & \textbf{0.09} & \textbf{79.48} & \textbf{27.65} \\
        \bottomrule
    \end{tabular}
    \vspace{-3mm}
    \caption{\textbf{Counting evaluation on 100 synthetic scenes.} 
    }
    \label{tab:eval_counting}
    \vspace{1em}

    \begin{tabular}{l r r r r}
        \toprule
        & NAE $\scriptscriptstyle\downarrow$ & SRE $\scriptscriptstyle\downarrow$ & MAE $\scriptscriptstyle\downarrow$ & sMAPE $\scriptscriptstyle\downarrow$ \\
        \midrule
        BMNet+~\cite{Shi22a} & 0.93 & 0.98 & 966.76 & 131.44 \\
        SAM+CLIP~\cite{Kirillov23,Radford21} & 0.94 & 0.99 & 980.33 & 124.31 \\ 
        CNN & 0.95 & 0.93 & 992.06 & 97.09  \\
        ViT+H & 0.94 & 0.93 & 979.29 & 91.45  \\
         \hline
        Human & 0.79 & 0.84 & 823.23 & 76.85 \\
        Human-Vote & 0.60 & 0.30 & 621.46 & 57.91 \\
        LlamaV 3.2 & 1.00 & 1.00 & 1037.5 & 190.48  \\
        \hline
        Ours (Color) & 0.57 & 0.27 & 607.98 & 74.33   \\
        Ours & \textbf{0.36} & \textbf{0.06} & \textbf{382.59} & \textbf{53.31} \\
        \bottomrule
    \end{tabular}
    \vspace{-3mm}
    \caption{\textbf{Counting evaluation on real-world scenes.}}
    \label{tab:eval_counting_real}
    \vspace{1em}

    \begin{tabular}{l c c c c}
        \toprule
        & MAE $\scriptscriptstyle\downarrow$ & RMSE $\scriptscriptstyle\downarrow$ & sMAPE $\scriptscriptstyle\downarrow$ & $R^2$ $\scriptscriptstyle\uparrow$\\ 
        \midrule
        \textit{DepthExtrapolated} & 0.36 & 0.38 & 77.43 & -6.04 \\
        \textit{DepthCorrected} &  0.10 & 0.12 & 34.80 & 0.28  \\
        Mean Estimator & 0.12 & 0.14 & 41.25 & 0.00 \\
         \hline
        Ours & \textbf{0.06} & \textbf{0.07} & \textbf{29.18} & \textbf{0.79} \\
        \bottomrule
    \end{tabular}
     \vspace{-3mm}
    \caption{\textbf{Occupancy ratio estimation.} We evaluate our method against three additional baselines that are tasked with prediction the occupancy ratio $\gamma$ from a depth map.}
    \label{tab:eval_volume}  
     \end{small} 
\end{table}

%% file: figs/additional.tex
\begin{figure}[t]
    \centering
    \begin{subfigure}{0.48\columnwidth}
        \centering
        \includegraphics[width=\linewidth]{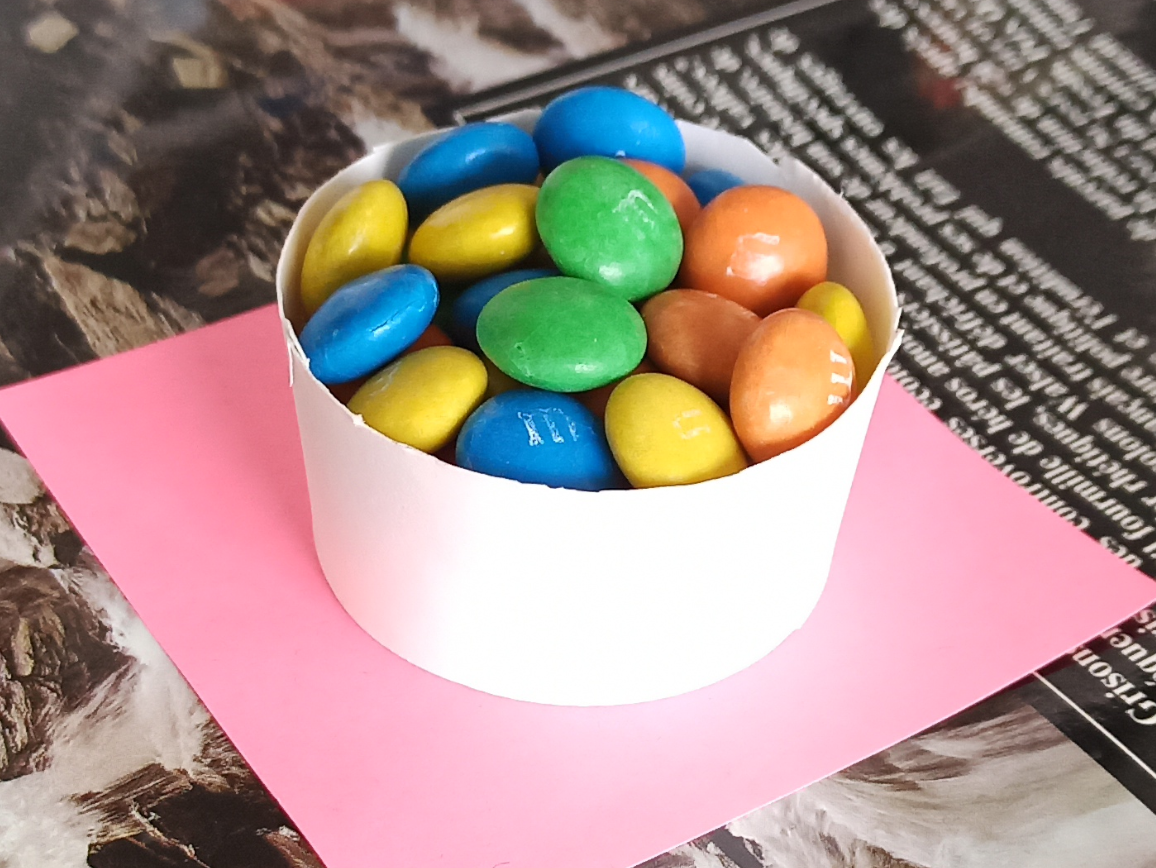} 
        \caption{$\mathcal{N}_{est} = 38$, $\mathcal{N}_{gt} = 36$}
    \end{subfigure}
    \hfill
    \begin{subfigure}{0.48\columnwidth}
        \centering
        \includegraphics[width=\linewidth]{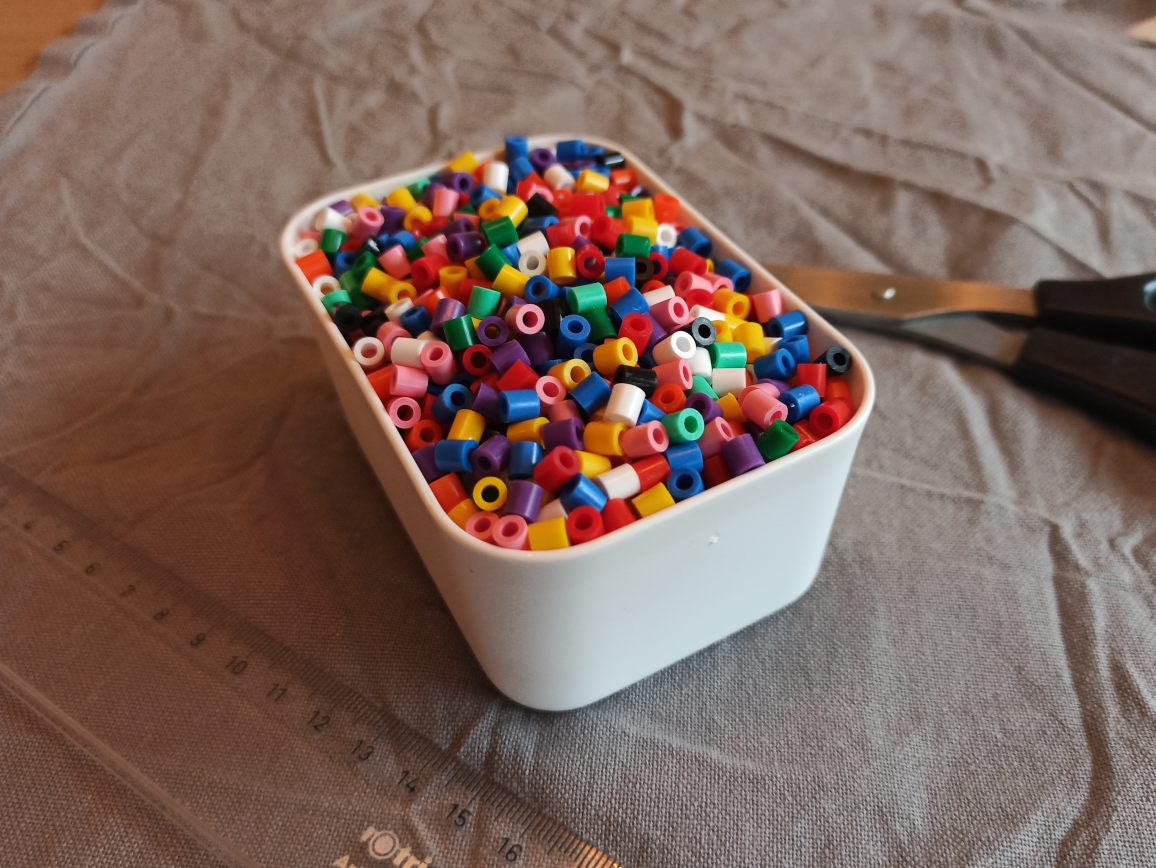} 
        \caption{$\mathcal{N}_{est} = 2133$, $\mathcal{N}_{gt} = 1830$}
    \end{subfigure}

    \vspace{0.5em} 

    \begin{subfigure}{0.48\columnwidth}
        \centering
        \includegraphics[width=\linewidth]{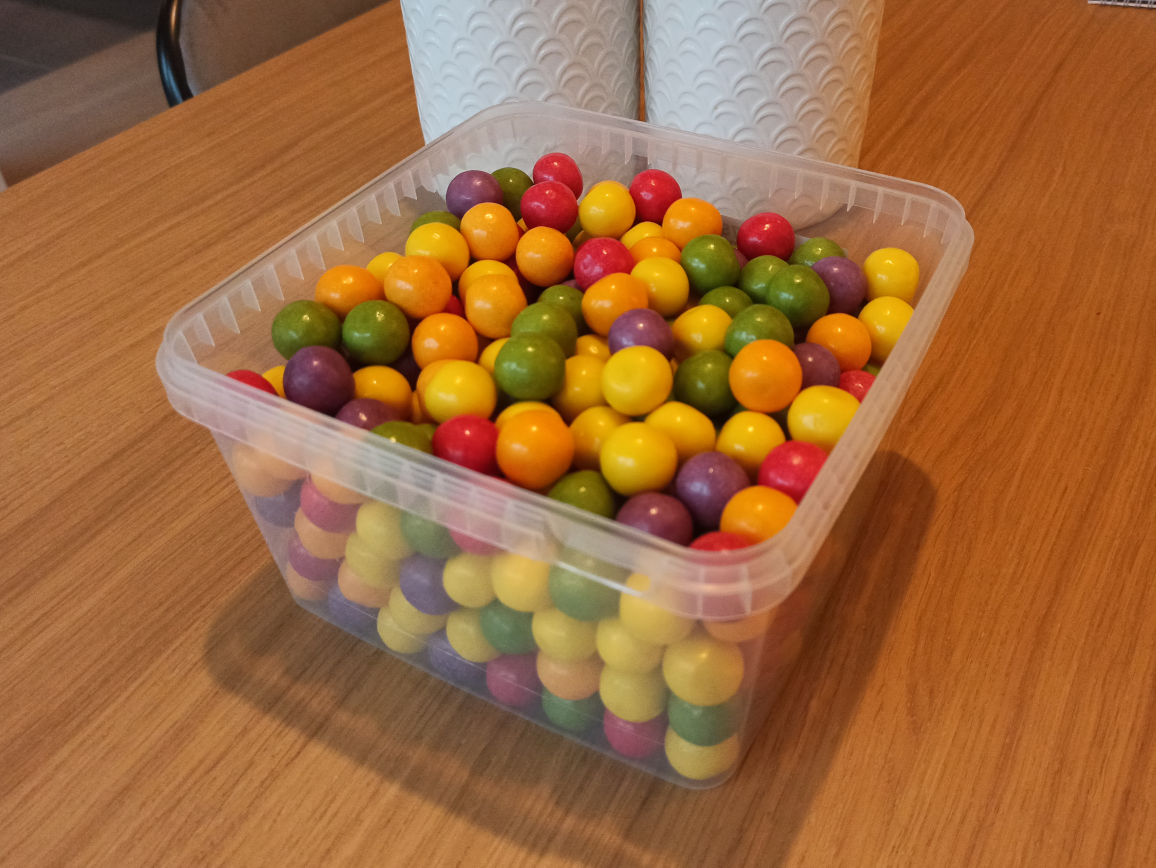} 
        \caption{$\mathcal{N}_{est} = 338$, $\mathcal{N}_{gt} = 397$}
    \end{subfigure}
    \hfill
    \begin{subfigure}{0.48\columnwidth}
        \centering
        \includegraphics[width=\linewidth]{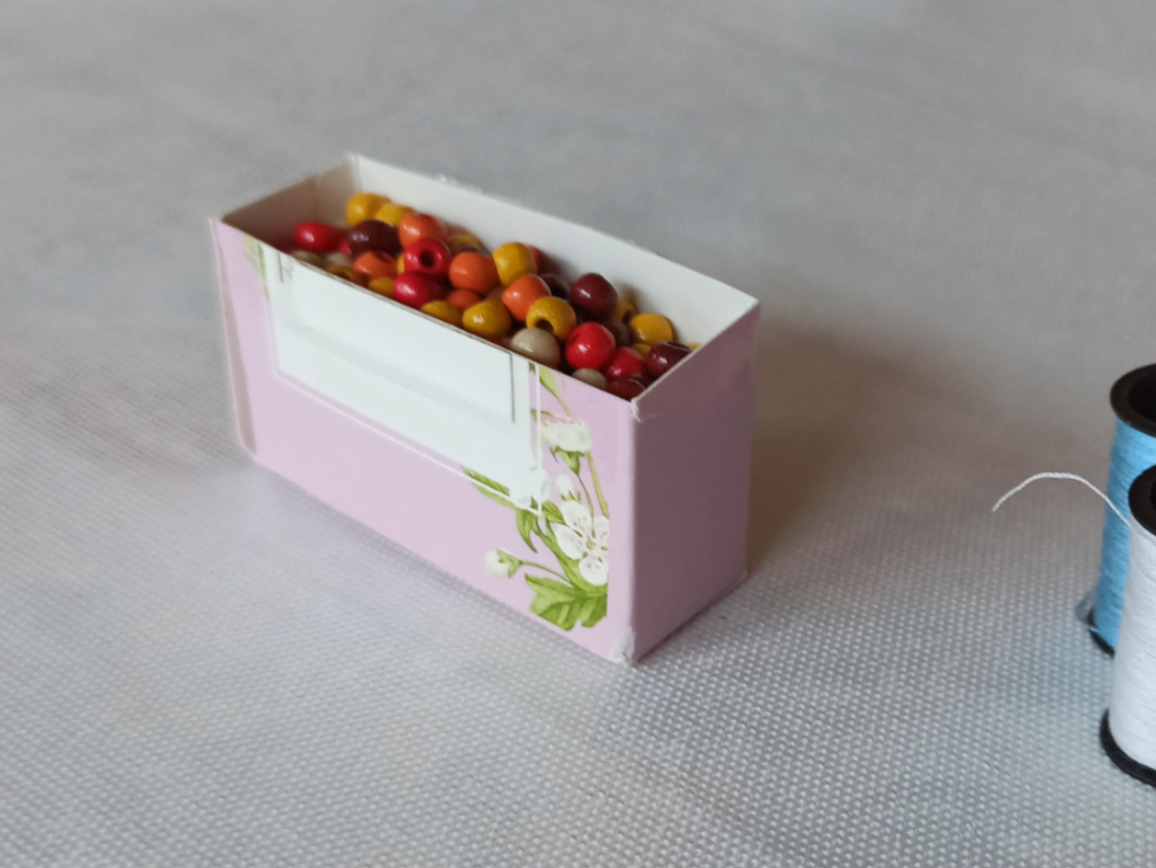} 
        \caption{$\mathcal{N}_{est} = 261$, $\mathcal{N}_{gt} = 300$}
    \end{subfigure}

    \caption{\textbf{Additional qualitative results.}}
    \label{fig:additional}
\end{figure}

%% file: figs/intermediate.tex
\begin{figure*}[t]
    \centering
    \begin{subfigure}{0.32\textwidth}
        \centering
        \begin{subfigure}{0.48\textwidth}
            \centering
            \includegraphics[width=\linewidth]{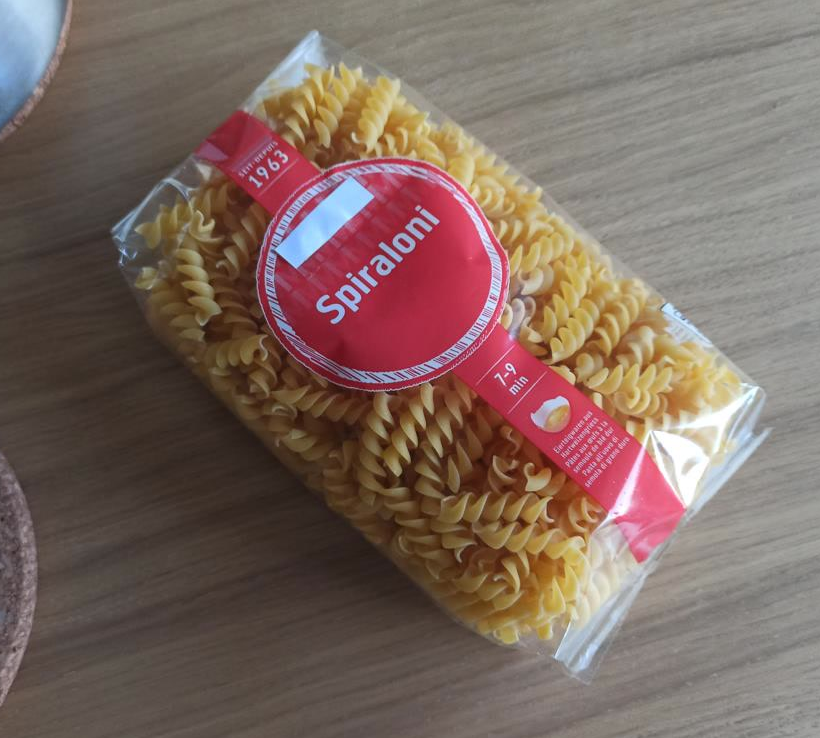} 
            \subcaption*{(a)}
        \end{subfigure}
        \hfill
        \begin{subfigure}{0.48\textwidth}
            \centering
            \includegraphics[width=\linewidth]{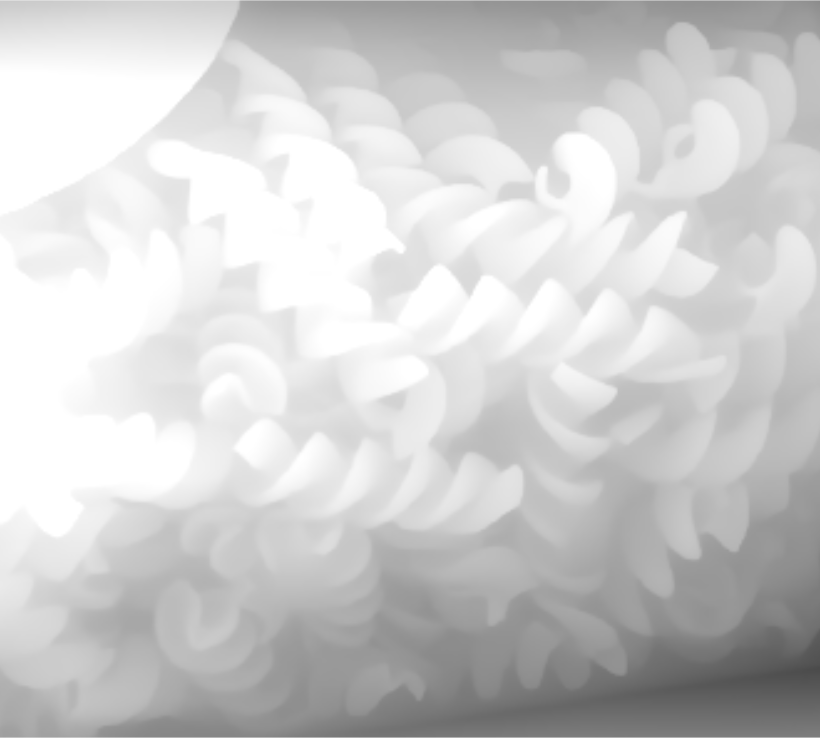}
            \subcaption*{(b)}
        \end{subfigure}

        \vspace{0.0em}

        \begin{subfigure}{0.48\textwidth}
            \centering
            \includegraphics[width=\linewidth]{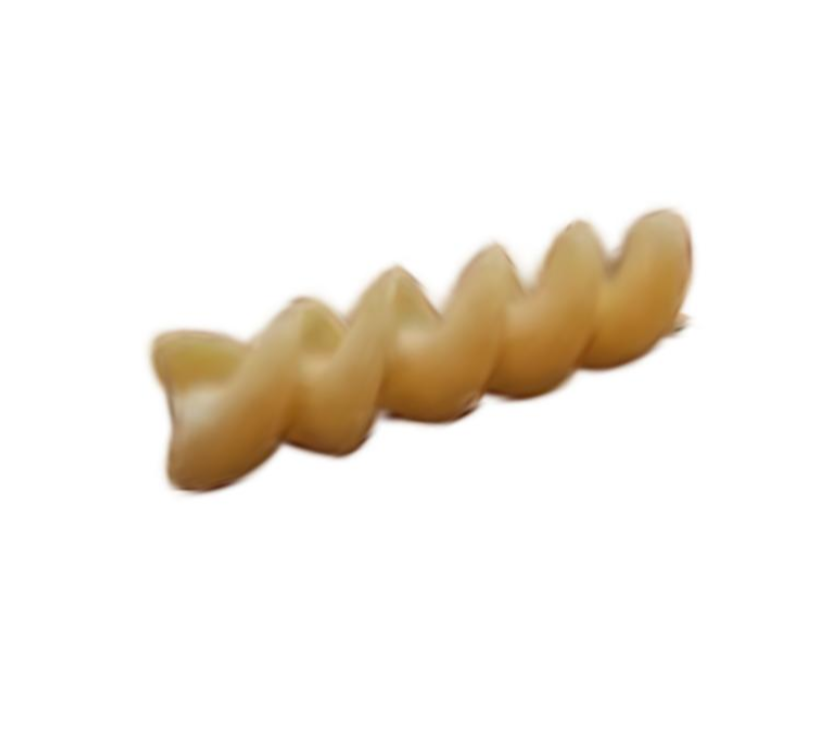}
            \subcaption*{(c)}
        \end{subfigure}
        \hfill
        \begin{subfigure}{0.48\textwidth}
            \centering
            \includegraphics[width=\linewidth]{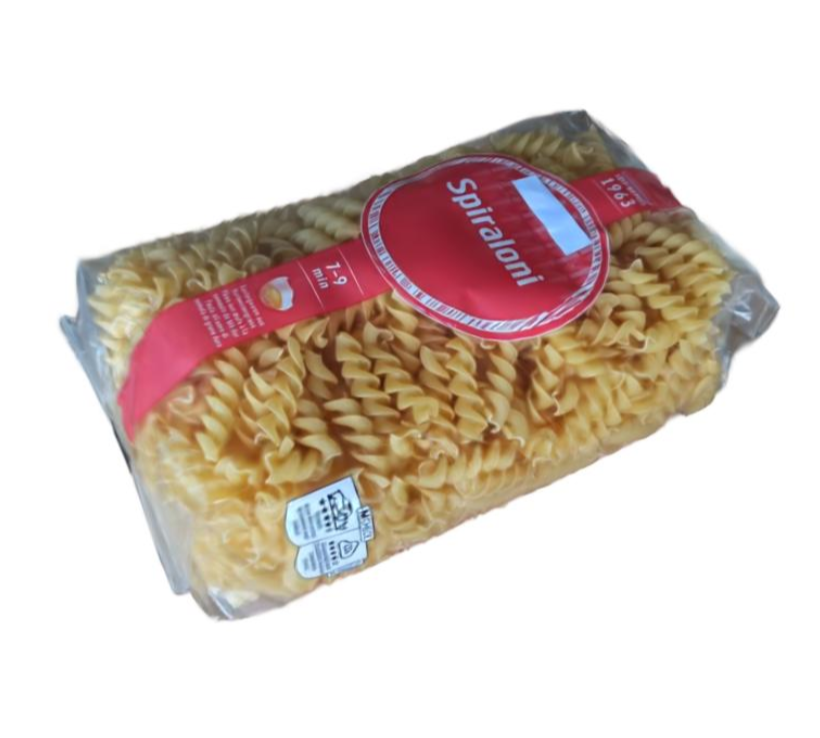}
            \subcaption*{(d)}
        \end{subfigure}
    \end{subfigure}
    \hfill
    \begin{subfigure}{0.32\textwidth}
        \centering
        \begin{subfigure}{0.48\textwidth}
            \centering
            \includegraphics[width=\linewidth]{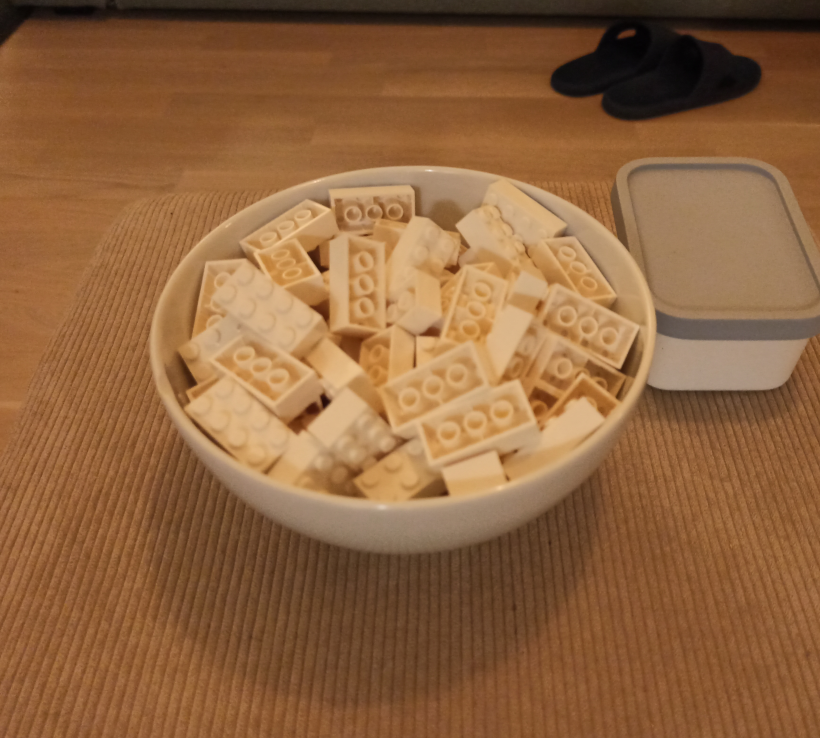}
            \subcaption*{(a)}
        \end{subfigure}
        \hfill
        \begin{subfigure}{0.48\textwidth}
            \centering
            \includegraphics[width=\linewidth]{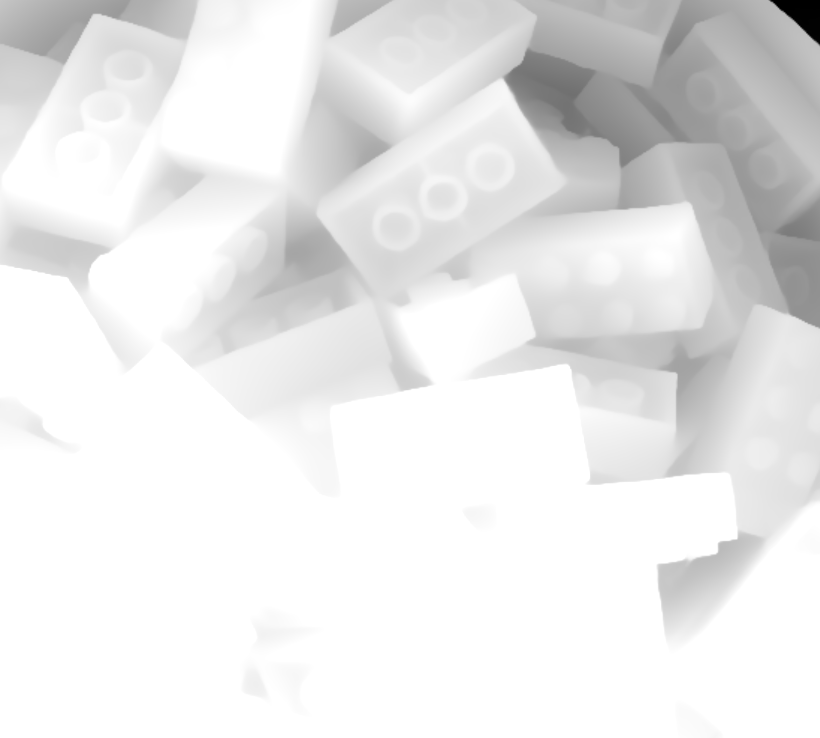}
            \subcaption*{(b)}
        \end{subfigure}

        \vspace{0.0em}

        \begin{subfigure}{0.48\textwidth}
            \centering
            \includegraphics[width=\linewidth]{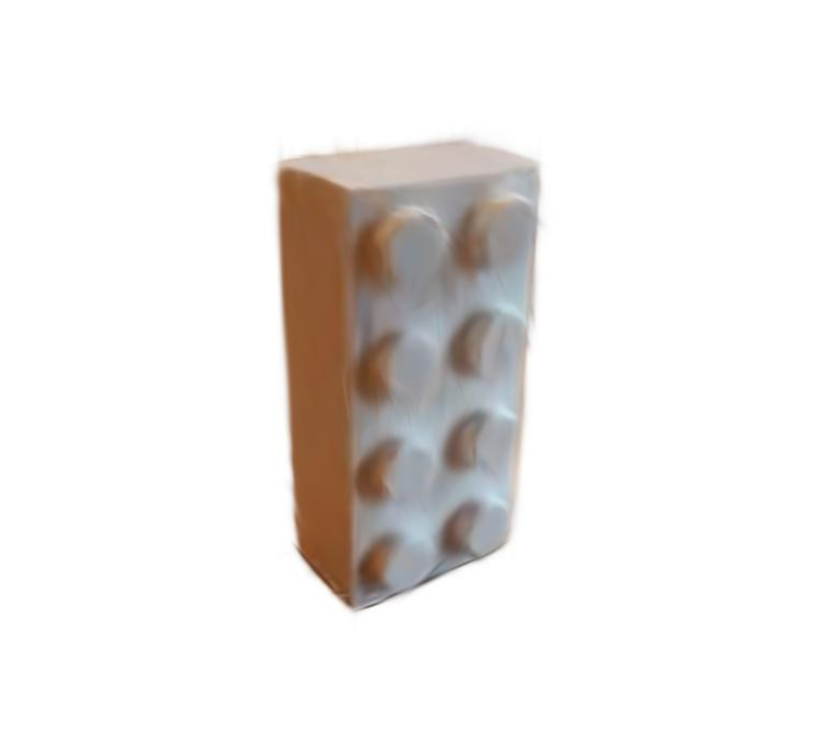}
            \subcaption*{(c)}
        \end{subfigure}
        \hfill
        \begin{subfigure}{0.48\textwidth}
            \centering
            \includegraphics[width=\linewidth]{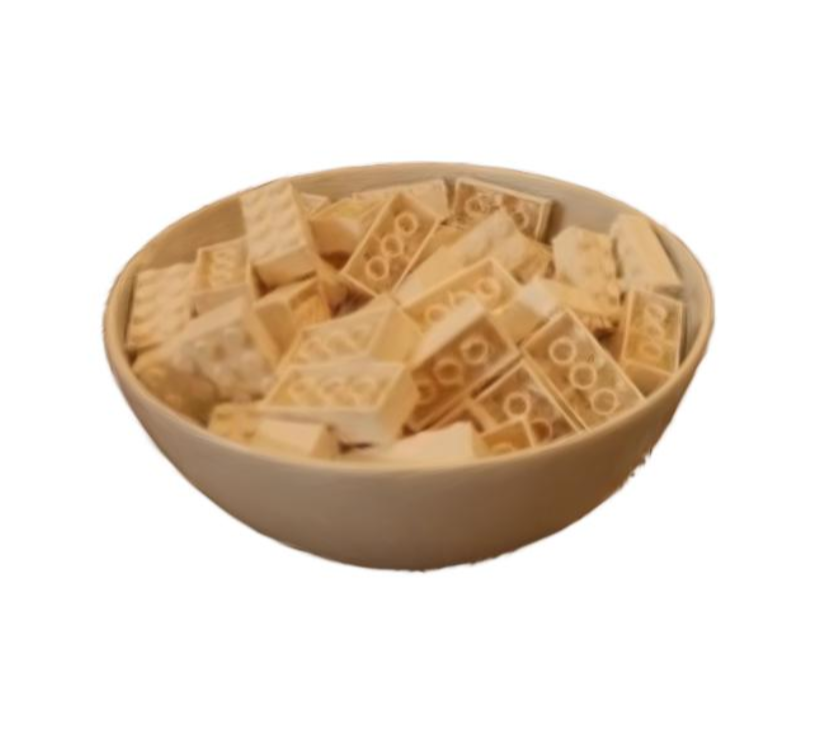}
            \subcaption*{(d)}
        \end{subfigure}
    \end{subfigure}
    \hfill
    \begin{subfigure}{0.32\textwidth}
        \centering
        \begin{subfigure}{0.48\textwidth}
            \centering
            \includegraphics[width=\linewidth]{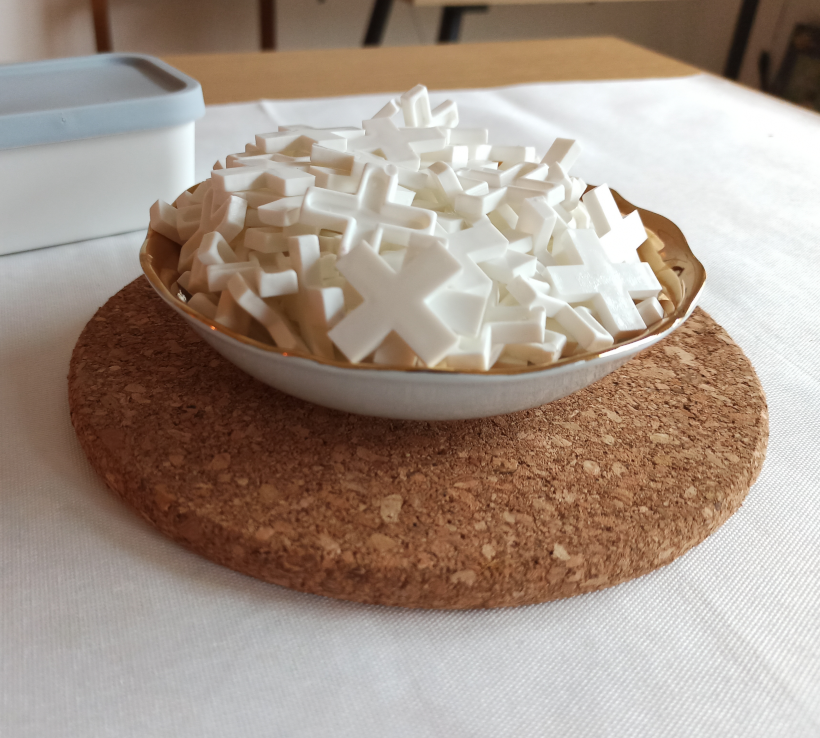}
            \subcaption*{(a)}
        \end{subfigure}
        \hfill
        \begin{subfigure}{0.48\textwidth}
            \centering
            \includegraphics[width=\linewidth]{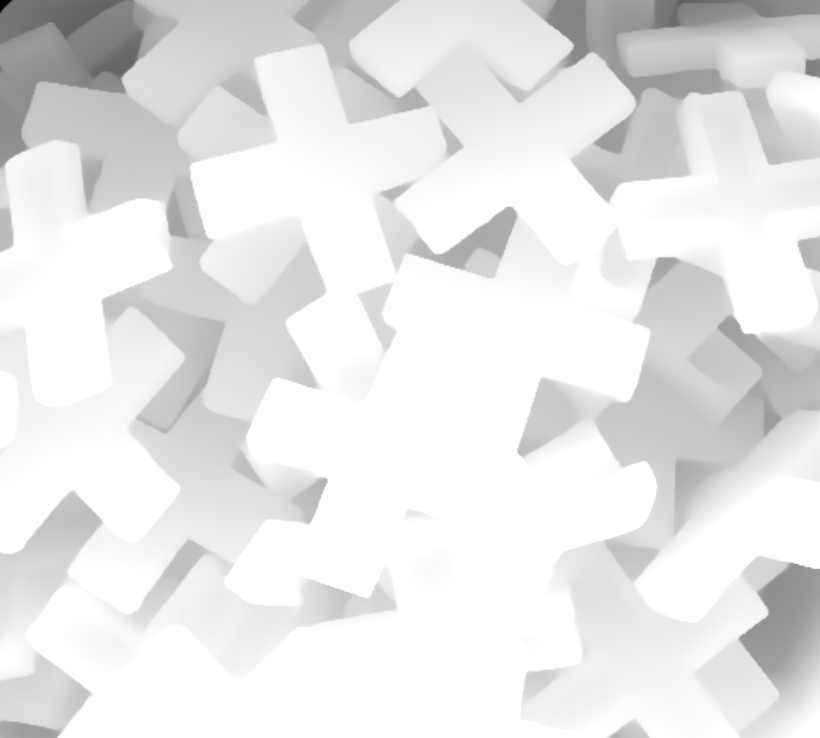}
            \subcaption*{(b)}
        \end{subfigure}

        \vspace{0.0em}

        \begin{subfigure}{0.48\textwidth}
            \centering
            \includegraphics[width=\linewidth]{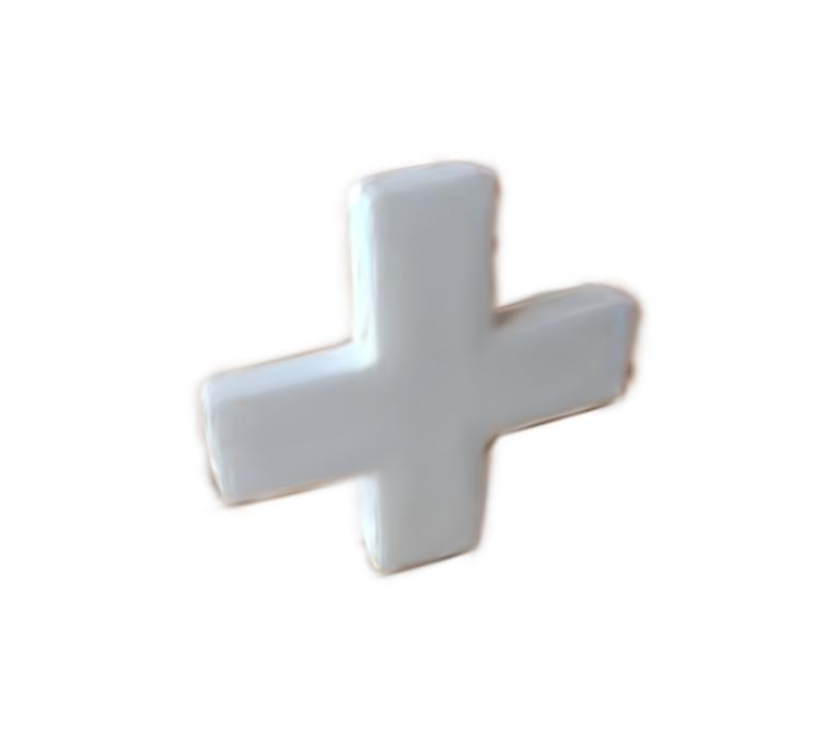}
            \subcaption*{(c)}
        \end{subfigure}
        \hfill
        \begin{subfigure}{0.48\textwidth}
            \centering
            \includegraphics[width=\linewidth]{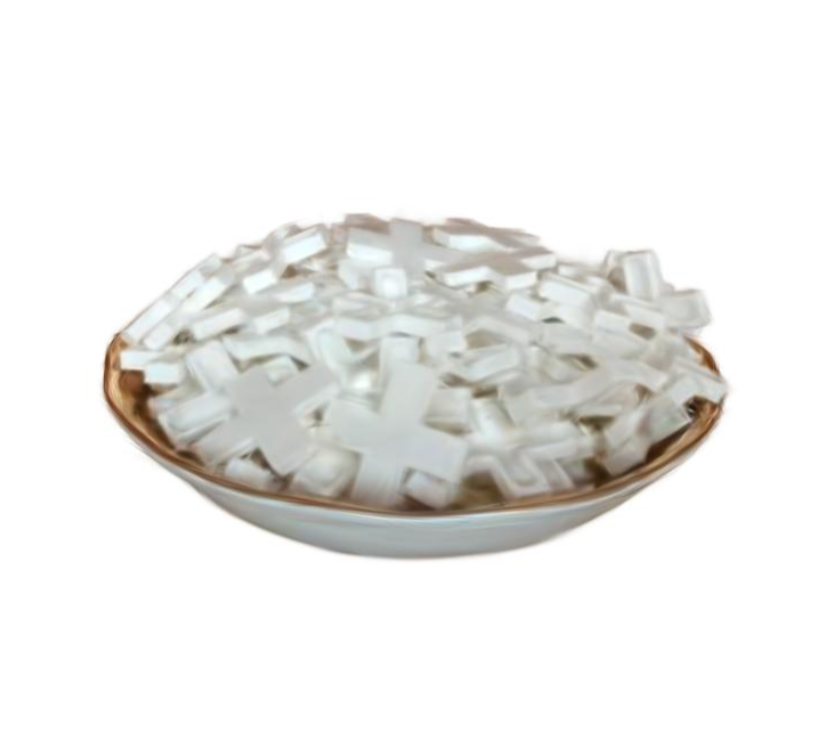}
            \subcaption*{(d)}
        \end{subfigure}
    \end{subfigure}

    \caption{\textbf{Intermediate results.} \cd{From images (a), we find the key viewpoint and compute its depth (b) to estimate the occupancy ratio $\gamma$. Using the unit shape previously reconstructed from images of a template (c), and the overall reconstruction (d), we deduce the final count. (\textit{Pasta}: $\gamma_{est}=30.5\%$, $N_{est}=509$, $N_{gt}=588$ . \textit{Bricks}: $\gamma_{est}=31.8\%$, $N_{est}=73$, $N_{gt}=100$ . \textit{Crosses}: $\gamma_{est}=29.6\%$, $N_{est}=88$, $N_{gt}=116$ )}}
    \label{fig:intermediate_results}
\end{figure*}

%% file: figs/ablation_table.tex
\begin{table}
    \centering
    \hspace{2em}

    \begin{tabular}{l r r r r}
        \toprule
        & NAE $\downarrow$ & SRE $\downarrow$ & MAE $\downarrow$ & sMAPE $\downarrow$ \\
        \midrule
        ($\mathcal{T}-$, $\mathcal{D}-$) & \textbf{0.22} & \textbf{0.09} & \textbf{79.48} & \textbf{27.65}  \\
        ($\mathcal{T}+$, $\mathcal{D}-$) & 0.28 & 0.11 & 100.12 & 30.93  \\
        ($\mathcal{T}+$, $\mathcal{D}+$) & 0.31 & 0.12 & 111.04 & 35.92 \\
        \bottomrule
    \end{tabular}
    \vspace{-0.5em}
    \caption{\textbf{Ablation study on 3D counting.} If ground-truth depth maps are used during training, it is indicated as $\mathcal{T}+$, and $\mathcal{T}-$ otherwise. Similarly, for evaluation purposes if ground-truth depth-maps are used during validation, we indicate it as $\mathcal{D}+$.
            }
    \label{tab:ablation_counting}
    \vspace{1em}

    \begin{tabular}{l c c c c}
        \toprule
        & MAE $\downarrow$ & RMSE $\downarrow$ & sMAPE $\downarrow$ & $R^2$ $\uparrow$\\ 
        \midrule
        ($\mathcal{T}-$, $\mathcal{D}-$) & \textbf{0.06} & \textbf{0.07} & \textbf{29.18} & \textbf{0.79} \\
        ($\mathcal{T}+$, $\mathcal{D}-$) & 0.08 & 0.11 & 32.01 & 0.52 \\
        ($\mathcal{T}+$, $\mathcal{D}+$) & 0.10 & 0.13 & 37.35 & 0.32 \\
        \bottomrule
    \end{tabular}
    \vspace{0em}
    \caption{\textbf{Ablation study on occupancy ratio estimation.} }
    \label{tab:ablation_volume}
\end{table}

%% file: figs/limitations.tex
\begin{figure}[h!]
    \centering
    \begin{adjustbox}{max width=\columnwidth}
    \begin{tabular}{c}
        \begin{subfigure}[b]{0.32\linewidth}
            \centering
            \includegraphics[width=\linewidth]{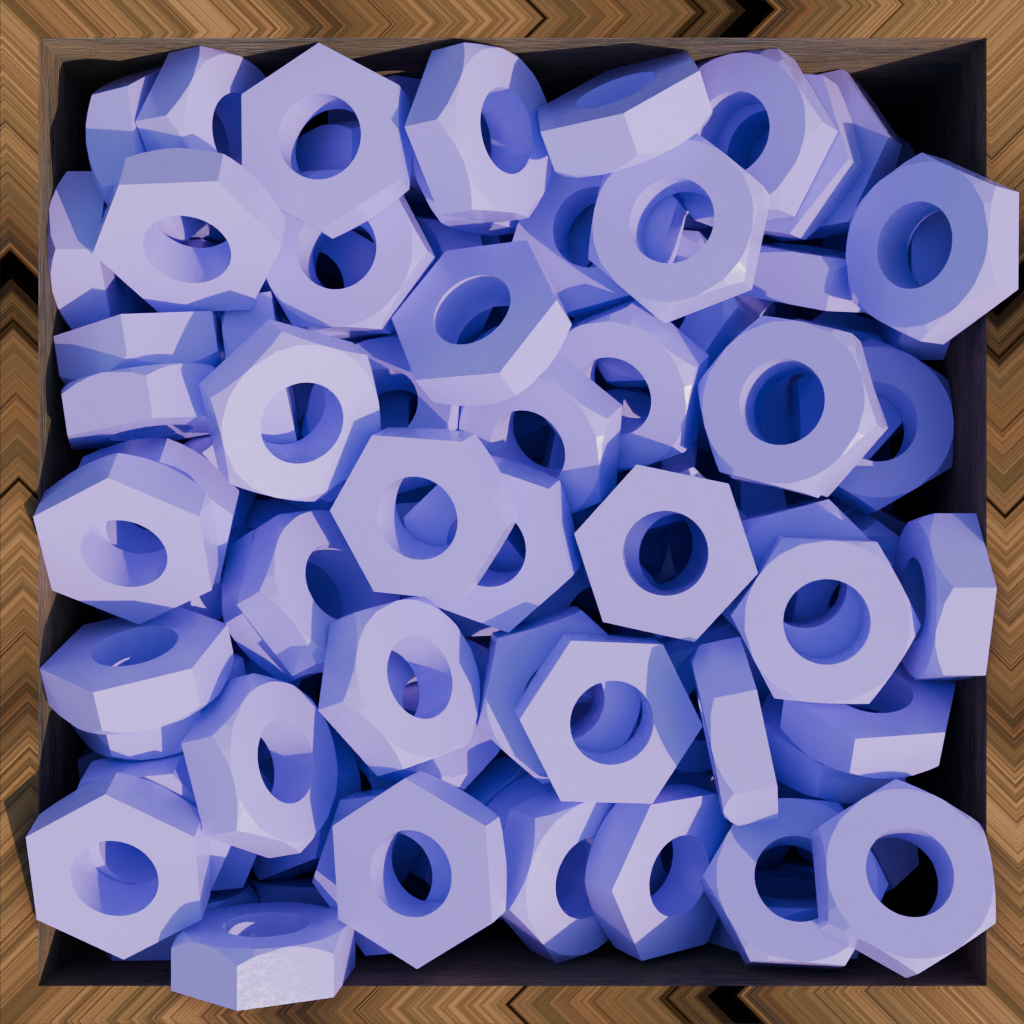}
            \caption{\scriptsize{Input image}}
        \end{subfigure}
        
        \begin{subfigure}[b]{0.32\linewidth}
            \centering
            \includegraphics[width=\linewidth]{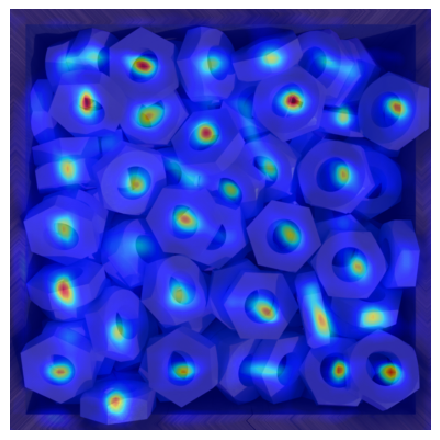}
            \caption{\scriptsize{BMNet+~\cite{Shi22a}}}
        \end{subfigure}
        
        \begin{subfigure}[b]{0.32\linewidth}
            \centering
            \includegraphics[width=\linewidth]{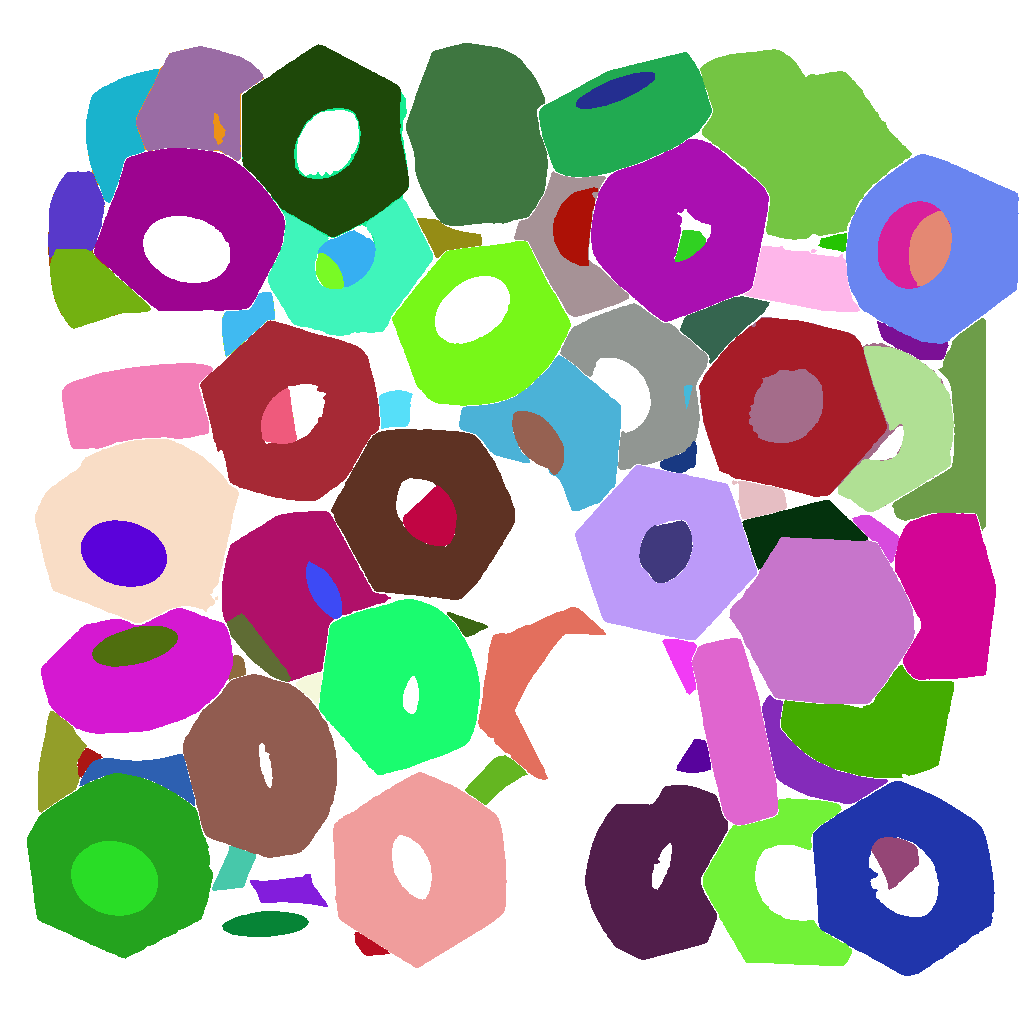}
            \caption{ \scriptsize{SAM\cite{Kirillov23}+CLIP\cite{Radford21}}}
        \end{subfigure}
    \end{tabular}
    \end{adjustbox}
    \vspace{-1em}
    \centering
    \caption{\textbf{Instance localization.} Previous methods also produce interpretable results, representing a promising direction for future work.}
    \label{fig:limitations}
\end{figure}

%% file: sec/5_conclusion.tex
\vspace{-1em}
\section{Conclusion}
\label{sec:conclusion}

In this paper, we introduced a novel method to count sets of stacked nearly-identical objects. In this scenario,  occlusions and irregular arrangements make accurate counting difficult. By decomposing the counting task into complementary subproblems---estimating independently the  3D volume of the stacks and the proportion of this volume actually occupied by objects---we were able to propose an effective solution that is easy to implement and far outperforms humans at this highly non-trivial task. 

Our experiments show that the performance of our approach can degrade with increased geometric complexity or visually complex scenes. In future work, we will therefore look into training the volume occupancy estimator to overcome these challenges. 
More generally, we believe our method and proposed datasets will open new applications and encourage future works centered on stacks of 3D objects, including 3D reconstruction, counting, or  3D scene understanding.

%% file: sec/X_suppl.tex
\clearpage
\setcounter{page}{1}
\maketitlesupplementary

\section{Dataset Details}
\label{sec:dataset_details}

\subsection{3DC-Real dataset.}
We capture 45 real scenes where the objects to count can be any stack of items that are at least partially visible. This includes stacked objects on a table or on the floor, objects in containers such as bowls or boxes, or objects still in their packaging. 

\parag{Cameras.}
We use a regular RGB smartphone camera to capture 30-60 pictures of the scene from various angles, forming a semisphere surrounding the objects and their container. These images are downscaled to approximately 600 pixels wide to reduce memory usage and facilitate the processing with COLMAP \cite{Schoenberger16a}. Additionally, we take a measurement of an arbitrary object within the scene, allowing us to scale the camera measurements and align the unit distance of the scene with a meter in the real world. 

Initially, we experimented with triangulation methods using two pairs of corresponding points across images. This would enable the calculation of a 3D distance and allow us to scale the scene. However, this approach proved to be unstable, as small inaccuracies in point matching led to significant variations in the scaling. Instead, we reconstructed the 3D scene with 3DGS \cite{Kerbl23} and measured the 3D distance directly within the reconstruction. This measurement allows us to rescale the scene to match the reference measurement. Note that the 3D point cloud generated by COLMAP \cite{Schoenberger16a}, which is used as an initialization by 3DGS, is also scaled accordingly.

\parag{Unit volume.}
For each scene, we require the unit volume of the object being counted. For complex shapes, we determine the unit volume $v$ by 3D reconstruction, similar to the method described in our main paper. For many common food items, such as kidney beans or corn, this information is readily available online. For other scenes, the volume can be approximated, for example in the case of the beads in \cref{fig:teaser}, by subtracting the volume of a cylinder from that of a sphere. 

\parag{Pre and Post-processing.}
Using this method, we capture 45 scenes consisting of various items in different environments. The scenes vary in complexity, from simple quasi-spherical objects in containers to more challenging configurations, such as complicated shapes still in packaging (e.g., in the \textit{pasta} scene). Some items are used in several scenes, in which case the container and location are modified to create a new setting.

\begin{figure*}[t]
    \centering
    \includegraphics[width=\linewidth]{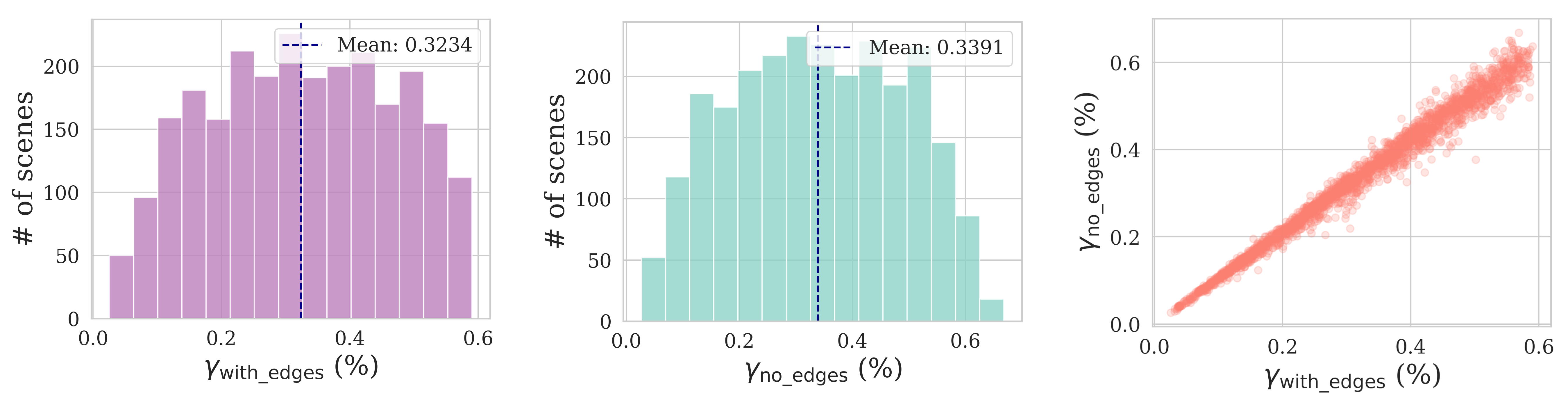}
    \caption{\textbf{Border effects.} Measuring the ground-truth $\gamma$ over the complete box or over only a smaller section makes little difference, indicating that our assumption of uniform $\gamma$ across the volume is justified.}
    \label{fig:supp_gamma_no_edges}
\end{figure*}

\vspace{1em}
\subsection{3DC-Synthetic Dataset.} 
To generate our large-scale synthetic dataset, we utilize Blender, a free and open-source 3D creation suite that supports Python scripting. This allows us to implement a fully automated generation pipeline, which is mainly composed of two steps: simulation and rendering.

\parag{Simulation.}
We drop batches of objects, arranged in a $4 \times 4 \times 5$ grid, into a box positioned at $(0, 0, 0.5)$. The box has a side length of $1$ and a thickness of $0.04$. Once the simulation converges, we check if the union of the objects intersects with an invisible cube placed directly on top of the box. If an intersection occurs, the simulation stops, and objects outside the box are deleted. If no intersection is detected, a new batch of objects is added, and the simulation is performed again.

We use the convex hull to compute collision between objects. Ideally, we would use the triangle mesh itself, however this becomes far too costly when physically simulating thousands of shapes with tens of thousands of triangles. We experimented with using convex hulls first, and then refining with additional frames using the triangle mesh, but this turned out to still be extremely costly and computationally very unstable, leading to objects being ejected outside the box due to the change in collision computation.

\parag{Rendering.}
For rendering, we use a texture randomly sampled from 3 possibilities for the box, five textures for the ground, and a random material for each model chosen from one of the following: a realistic grey metal texture, a red metallic texture, or a plastic material with a randomly selected color. 

We always render the first view directly above the box, looking downwards, which we call the \textit{nadir} view. For the validation dataset, we also generate 29 additional views on the unit sphere, each observing the box from different angles. The rendering is performed using Blender's Cycles rendering engine. Additionally, we generate ground-truth depth maps and masks that separate the ground, box, and objects in the images.

\parag{Pre and Post-processing.}
In addition to the simulation and rendering steps, we perform pre-processing to filter out unsuitable meshes, such as those with multiple connected components or excessive size. Since the physical simulation can sometimes be unstable or fail, we also remove a small fraction of results in post-processing. This includes cases where the unit volume is too small or where too few objects remain in the box in the final frame.

Finally, we export the calibrated camera parameters in a format compatible with nerfstudio \cite{Tancik23}. Since these cameras are not produced by COLMAP, they do not include a 3D point cloud that 3DGS can use as initialization. This poses a challenge, as a fully random initialization may generate distant Gaussian points outside the cameras' range, which are not removed and interfere with the volume estimation. To address this, we generate a set of 100 grey points within the unit cube, centered at $(0, 0, 0.5)$. This simple initialization proves sufficient to quickly produce a faithful 3D reconstruction and resolves the aforementioned issue.

\section{Additional comparison distinguishing visible and invisible objects.}

\begin{figure}[t]
    \centering
    \hfill
    \begin{minipage}{0.45\columnwidth}
        \includegraphics[width=\linewidth]{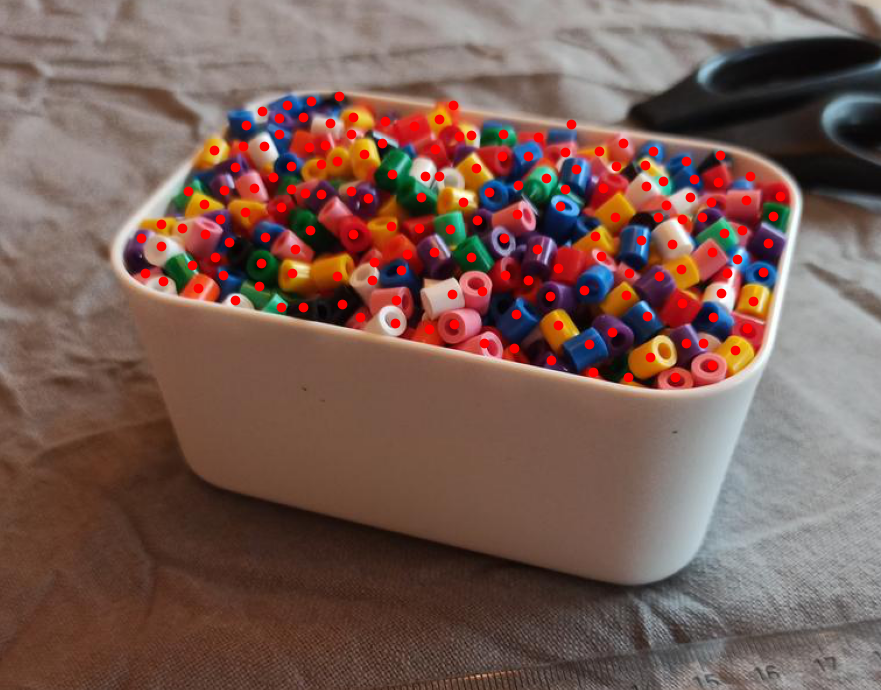}
    \end{minipage}
    \hfill
    \begin{minipage}{0.45\columnwidth}
        \includegraphics[width=\linewidth]{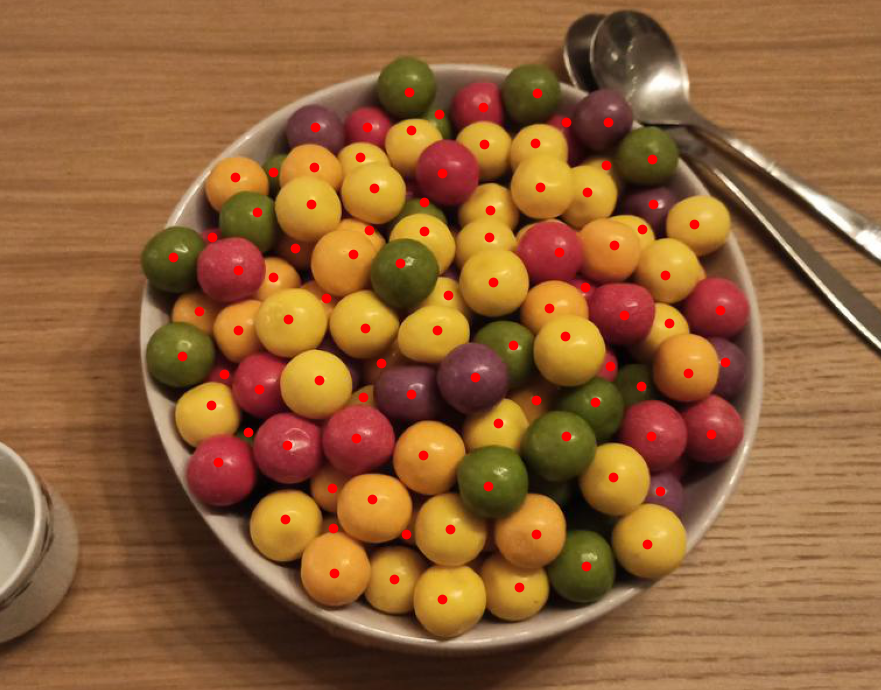}
    \end{minipage} 
    \hfill \vspace{-0.5em}
    \caption{\textbf{Localization of visible objects.} 
    Left: $\mathcal{N}_{vis}=168$, Right: $\mathcal{N}_{vis}=96$}
    \label{fig:localization}
    \vspace{-0.8em}
\end{figure}

Our method is, to the best of our knowledge, proposing the first solution for this task. The methods used as comparison in \ref{sec:experiments} are initially designed to count only visible objects, and perform poorly on our dataset. In this section, we distinguish between visible and invisible objects in stacks and evaluate baseline methods against both counts. To perform this evaluation, we manually annotate the locations of visible objects in real scenes and provide the visible count and locations in our released datasets. Two examples are displayed in~\cref{fig:localization}.

The results of this experiment can be found in~\cref{tab:eval_visible}. While the numbers for the 2D counting methods improve, it remains hard for them to distinguish similar objects clumped together. As a result, their performance re,ains much lower on these challenging scenes than on traditional 2D counting benchmarks.

\begin{table}
    \centering
    \begin{small} 
    \begin{tabular}{l r r r}
        \toprule
        & NAE $\scriptscriptstyle\downarrow$ & SRE $\scriptscriptstyle\downarrow$ & sMAPE $\scriptscriptstyle\downarrow$ \\
        \midrule
        \textit{Visible objects only} \\
        BMNet+ [21] & \textbf{0.51} & \textbf{0.65} & 51.28 \\
        SAM+CLIP [7,16] & 0.57 & 0.81 & \textbf{50.84} \\
        \hline
        \textit{All objects} \\
        BMNet+ [21] & 0.93 & 0.98  & 131.44 \\
        SAM+CLIP [7,16] & 0.94 & 0.99  & 124.31 \\ 
        Ours & \textbf{0.36} & \textbf{0.06} & \textbf{53.31} \\
        \bottomrule
    \end{tabular}
    \vspace{-2mm}
    \caption{\textbf{Counting visible and invisble objects separately.}
    }
    \vspace{-1.5em}
    \label{tab:eval_visible}
\end{small} 
\end{table}

\section{Evaluation metrics}

In this section, we provide the exact formula behind the metrics used in~\cref{sec:experiments}. As explained, the NAE and SRE are defined as:
\begin{align*}
\text{NAE} &= \frac{\sum_{i=1}^n |y_i - \hat{y}_i|}{\sum_{i=1}^n y_i}, \hspace{1em}
\text{SRE} &= \frac{\sum_{i=1}^n (y_i - \hat{y}_i)^2}{\sum_{i=1}^n y_i^2} \; .
\end{align*}
We also report the Symmetric Mean Absolute Percentage Error (sMAPE), which can be considered a normalized percentage error. It is expressed as follows:
\[
\text{sMAPE} = \frac{100\%}{n} \sum_{i=1}^n \frac{|y_i - \hat{y}_i|}{(y_i + \hat{y}_i) / 2},
\]
The formula of sMAPE ensures that errors are scaled symmetrically between the prediction and ground truth counts. Finally, the coefficient of determination, \( R^2 \), measures the proportion of variance in the ground-truth occupancy ratio $\gamma$ explained by our predictions
\[
R^2 = 1 - \frac{\sum_{i=1}^n (y_i - \hat{y}_i)^2}{\sum_{i=1}^n (y_i - \overline{y})^2},
\]
where \( \overline{y} \) is the mean of the ground truth counts. High values of \( R^2 \) indicate strong agreement between predictions and ground-truth values.

\begin{figure*}[t]
    \centering
    \includegraphics[width=\linewidth]{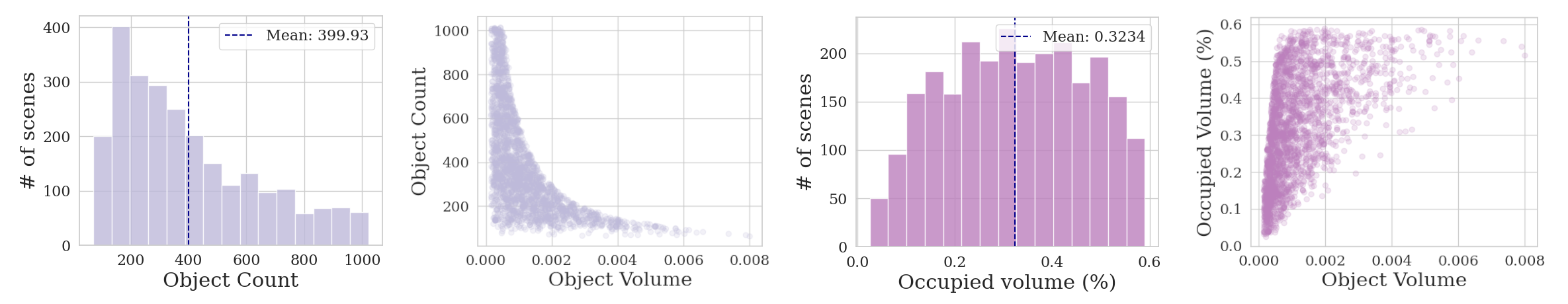}
    \caption{\textbf{Dataset statistics.} The histograms represent the distributions of object count and occupancy ratio, respectively, and each bar plots the number of scenes in a given bin.  In scatter plots, each point represents a physically simulated 3D scene. In particular, the occupancy ratio $\gamma$ spans a large range between $1\%$ and 65$\%$}
    \label{fig:dataset_stats}
\end{figure*}

\section{Additional Discussion on Border Effects}

In our work, we assume that the occupancy ratio, \(\gamma\), is approximately uniform throughout the container. This assumption generally holds as the number of stacked objects increases. However, it neglects the influence of container borders, where objects tend to occupy less volume due to the boundary.

To quantitatively evaluate the impact of border effects and verify the validity of our uniform \(\gamma\) assumption, we analyze the ground-truth volume ratio in two distinct ways using our large-scale synthetic dataset. First, we compute \(\gamma_{\text{with\_edges}}\) for the entire unit box, as described in the main paper. Additionally, we compute \(\gamma_{\text{no\_edges}}\) by measuring the volume ratio in a smaller sub-box of side length 0.5, centered within the unit box. Intuitively, the difference between \(\gamma_{\text{with\_edges}}\) and \(\gamma_{\text{no\_edges}}\) reflects the influence of border effects, allowing us to evaluate whether this assumption is justifiable.

\Cref{fig:supp_gamma_no_edges} presents two histograms comparing the distributions of \(\gamma_{\text{with\_edges}}\) and \(\gamma_{\text{no\_edges}}\). The results indicate that both metrics follow highly similar distributions, with their mean values differing by less than 5\%. Notably, the mean value of \(\gamma_{\text{no\_edges}}\) is slightly higher than that of \(\gamma_{\text{with\_edges}}\), consistent with the intuition that density decreases near borders.

To further investigate the relationship between these two values, we provide a scatter plot of \(\gamma_{\text{with\_edges}}\) versus \(\gamma_{\text{no\_edges}}\) in \cref{fig:supp_gamma_no_edges}. The plot demonstrates a strong correlation between the two measures, particularly for objects with small volume ratios. For objects with high values of both \(\gamma_{\text{with\_edges}}\) and \(\gamma_{\text{no\_edges}}\), minor discrepancies are observed. These differences can be attributed to the relatively large size of these objects compared to the measurement box, which introduces noise in the estimation of \(\gamma\).

Overall, these analyses confirm that \(\gamma_{\text{with\_edges}}\) and \(\gamma_{\text{no\_edges}}\) are highly consistent and can be used interchangeably without significant loss of accuracy. In our experiments, we rely on \(\gamma_{\text{with\_edges}}\) to train our occupancy ratio estimation network.

\section{Implementation details}
\label{sec:implementation_details}

We use the nerfstudio library~\cite{Tancik23} for 3D reconstruction, specifically the \textit{splatfacto} method built on top of the gsplat library~\cite{Ye24a}. We thank the contributors of all the aforementioned libraries.

Our pipeline also uses pretrained models for depth estimation and mask generation. We employ the \textit{vitl} model from Depth Anything v2~\cite{Yang24c} for depth estimation and the \textit{sam2.1\_hiera\_large} model from SAM2~\cite{Ravi24a} for mask generation. These state-of-the-art models ensure high-quality and robust outputs across diverse scenes.

The dataset is generated using CPUs only, greatly reducing its production cost and environmental impact. Other operations are fairly light and performed locally on a 4080 Mobile GPU, taking up only a few gigabytes of VRAM and being completed in a couple minutes.

\section{Architecture details}
\label{sec:architecture_details}

Our architecture utilizes a DinoV2~\cite{Oquab23} encoder model that produces pixel-aligned features. Since DinoV2 downscales the input image by 14, we feed it an image of size 448 x 448 to produce a 32 x 32 x 768 feature image. Specifically, we use the pretrained weights of the \textit{dinov2\_vitb14} model and freeze them during all subsequent learning.

To predict a scalar value from the \(32 \times 32 \times 768\) feature image produced by DinoV2, we employ a series of convolutional layers to progressively reduce both the spatial dimensions and the number of channels. The convolutional layers successively reduce the channel dimension from the initial 768 down to 512, 256, 128, and finally 64. Concurrently, the spatial dimensions of the feature map are reduced from \(32 \times 32\) to \(16 \times 16\), \(8 \times 8\), \(4 \times 4\), and ultimately \(2 \times 2\).

Following this, an adaptive average pooling layer compresses the spatial dimensions to a single pixel while preserving the 64-channel depth. The resulting \(1 \times 1 \times 64\) tensor is passed through a fully connected linear layer to map it to a scalar output. Finally, a sigmoid activation function is applied to produce the final prediction in the $[0,1]$ range.